\documentclass[letterpaper]{article} 
\usepackage{aaai23}  
\usepackage{times}  
\usepackage{helvet}  
\usepackage{courier}  
\usepackage[hyphens]{url}  
\usepackage{graphicx} 
\usepackage{enumitem}
\usepackage{amsmath}
\usepackage{amssymb}
\usepackage{multirow}
\usepackage{booktabs}
\usepackage{array}
\usepackage{bm}
\usepackage{xcolor,colortbl}
\usepackage{bbding}
\usepackage{pifont}
\usepackage{natbib}  
\usepackage{caption} 
\usepackage{algorithm}
\usepackage{algorithmic}

\usepackage{newfloat}
\usepackage{listings}
\DeclareCaptionStyle{ruled}{labelfont=normalfont,labelsep=colon,strut=off} 
\floatstyle{ruled}
\newfloat{listing}{tb}{lst}{}
\floatname{listing}{Listing}
\definecolor{color4}{rgb}{0.94,0.94,1}
\title{Intriguing Findings of Frequency Selection for Image Deblurring}
\author{
    Xintian Mao\textsuperscript{\rm 1}, Yiming Liu\textsuperscript{\rm 1}, Fengze Liu\textsuperscript{\rm 2}, Qingli Li\textsuperscript{\rm 1}, Wei Shen\textsuperscript{\rm 3}, Yan Wang\textsuperscript{\rm 1}\thanks{Corresponding Author: \tt ywang@cee.ecnu.edu.cn.}\\
}
\affiliations{
    \textsuperscript{\rm 1}Shanghai Key Laboratory of Multidimensional Information Processing, East China Normal University\\
    \textsuperscript{\rm 2}Department of Computer Science, the Johns Hopkins University\\
    \textsuperscript{\rm 3}MoE Key Lab of Artificial Intelligence, AI Institute, Shanghai Jiao Tong University\\
    \tt \small \{52265904010,52205904009\}@stu.ecnu.edu.cn, fliu23@jhu.edu, qlli@cs.ecnu.edu.cn, wei.shen@sjtu.edu.cn, ywang@cee.ecnu.edu.cn \\
}

\usepackage{bibentry}

\begin{document}

\maketitle

\begin{abstract}
Blur was naturally analyzed in the frequency domain, by estimating the latent sharp image and the blur kernel given a blurry image. Recent progress on image deblurring always designs end-to-end architectures and aims at learning the difference between blurry and sharp image pairs from pixel-level, which inevitably overlooks the importance of blur kernels. This paper reveals an intriguing phenomenon that simply applying \emph{ReLU} operation on the \emph{frequency} domain of a blur image followed by inverse Fourier transform, \emph{i.e.}, frequency selection, provides faithful information about the \emph{blur pattern} (\emph{e.g.}, the blur direction and blur level, implicitly shows the kernel pattern). Based on this observation, we attempt to leverage kernel-level information for image deblurring networks by inserting Fourier transform, ReLU operation, and inverse Fourier transform to the standard ResBlock. 1 $\times$ 1 convolution is further added to let the network modulate flexible thresholds for frequency selection. We term our newly built block as {Res FFT-ReLU Block}, which takes advantages of both kernel-level and pixel-level features via learning frequency-spatial dual-domain representations. 
Extensive experiments are conducted to acquire a thorough analysis on the insights of the method. Moreover, after plugging the proposed block into NAFNet, we can achieve 33.85 dB in PSNR on GoPro dataset. Our method noticeably improves backbone architectures without introducing many parameters, while maintaining low computational complexity. Code is available at \url{https://github.com/DeepMed-Lab/DeepRFT-AAAI2023}.
\end{abstract}

\section{Introduction}
\label{sec:intro}
Image deblurring aims at removing blurring artifacts to recover sharp images \cite{Cho2021rethinking}. The blurring of an image can be caused by many factors, \textit{e.g.}, camera shake, objects movement, out-of-focus optics, etc. The blurry image leads to visually low quality and hampers subsequent high-level vision tasks, ranging from security, medical imaging to object recognition \cite{Chen2020one}. 

\begin{figure}[t]
\begin{center}
    \includegraphics[width=0.8\linewidth]{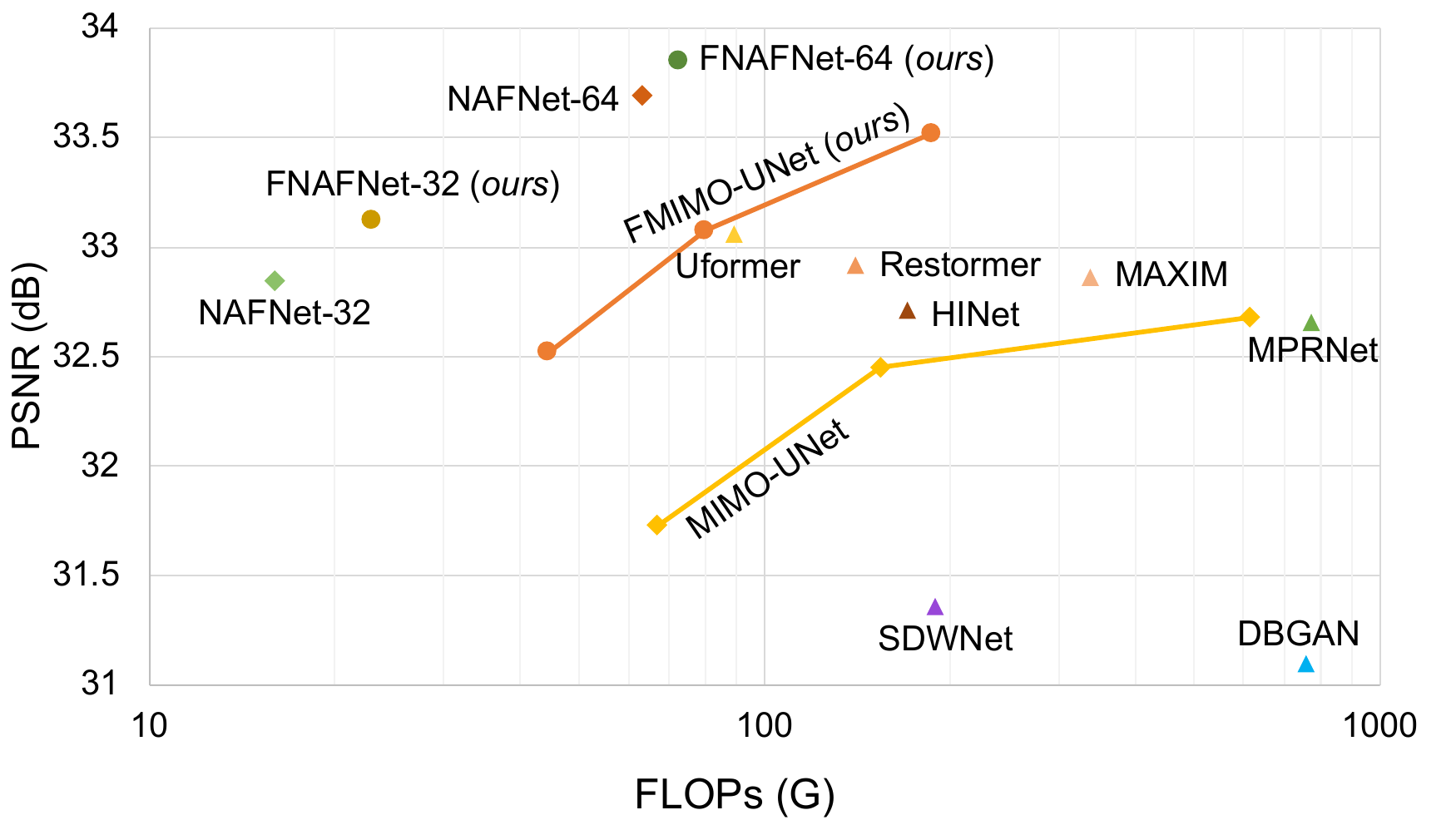}
\end{center}
\caption{PNSR vs. computational cost on the GoPro dataset \cite{Nah2017deep}. Our method performs much better than baseline methods: MIMO-UNet \cite{Cho2021rethinking} and NAFNet \cite{Chen2022simple}, and other state-of-the-arts. } 
\label{fig:glance}
\end{figure}

Image deblurring by frequency domain operations was very popular decades ago, based on a simple assumption that image blur may be due to the Point Spread Function (PSF) of the sensor, sensor motion, and other reasons \cite{Benham1997digital}.  
Thus, the motion blurred image is expressed by convolving a latent sharp image with the PSF, which can be most easily instantiated in the frequency domain \cite{Chakrabarti2010analyzing,Xu2013unnatural,Hu2014deblurring,Pan2016blind}. Here we take a simple example and assume that the blur kernel does not vary spatially:  
$B(\omega)=F(\omega)G(\omega)+N(\omega)$,
where $F(\omega)$, $G(\omega)$, $B(\omega)$ and $N(\omega)$ are Fourier transforms of the sharp image, the blur kernel (PSF), the blurry image and the sensor noise, respectively, and $\omega\in[-\pi,\pi]^2$.  This gives us an important cue that frequency domain offers us abundant information on image deblurring tasks which should not be overlooked.

Deep networks are popular for their end-to-end learning ability. The field of image deblurring has made significant advances riding on the wave of deep networks. DeepDeblur \cite{Nah2017deep}, pioneers the technique of end-to-end trainable methods, directly mapping a blurry image to its paired sharp image by a {C}onvolutional {N}eural {N}etwork (CNN). It designs a multi-scale architecture, and uses a modified residual network structure \cite{He2016deep} called \textit{ResBlock} (see Fig.~\ref{fig:blockcompare} (a)) to focus on learning the difference between blurry and sharp image pairs.
Thereafter, end-to-end learning strategy with ResBlock is proven to be effective in image deblurring, and becomes a mainstream approach in recent years \cite{Tao2018scale,Zhang2019deep,HongyunGao2019DynamicSD,HongguangZhang2019DeepSH,Park2020multi,YuanYuan2020EfficientDS,Zamir2021multi,Chen2021hinet,Zou2021sdwnet,KuldeepPurohit2021SpatiallyAdaptiveIR,Cho2021rethinking,Chen2022simple}. But, these methods overlook the importance of blur kernels.

In this paper, we reveal an intriguing phenomenon that taking the {inverse Fourier transform} on \emph{frequency selection} (\emph{e.g.}, ReLU on the frequency domain) of a blurry image acts as learning \emph{blur pattern} from the blurry image, indicating the blur direction and blur level, implicitly showing the \emph{kernel} pattern. The faithful blur kernel information provided by such operations motivates us to insert Fourier transform, ReLU, and inverse Fourier transform to the standard ResBlock to take advantages of both \emph{kernel}-level and \emph{pixel}-level features via fusing frequency-spatial dual-domain representations. Furthermore, we investigate ReLU on the frequency domain from a new perspective, and experiments show that setting different thresholds instead of $0+j0$ ($0+j0$ is for ReLU) for frequency selection give different deblurring results. With this new viewpoint, we find that adding convolution after Fourier transform helps
the network modulate flexible thresholds for selecting frequencies and can further promote image deblurring performance. To sum up, we propose a new, efficient and \emph{plug-and-play} ResBlock, termed as Residual (Res) Fast Fourier Transform (FFT)-ReLU Block, to replace standard ResBlock. Our Res FFT-ReLU Block inserts a FFT-ReLU stream, consisting of 4 simple operations: 2D real FFT, 1$\times$1 convolutions, ReLU and inverse 2D real FFT, into ResBlock.

The effectiveness of the Res FFT-ReLU Block is compared and verified by plugging in different architectures on three datasets: GoPro \cite{Nah2017deep}, HIDE \cite{Shen2019human} and RealBlur \cite{Rim2020real} datasets. Substantial ablation studies are conducted to explore insights of the  FFT-ReLU stream. It is worth mentioning that after plugging the proposed stream to MIMO-UNet+ \cite{Cho2021rethinking} and NAFNet \cite{Chen2022simple}, the new models, which are termed as FMIMO-UNet+ and FNAFNet, can achieve 33.52 dB and 33.85 dB respectively in terms of PSNR on GoPro dataset. Our method noticeably improves backbone architectures without introducing too many parameters, while maintaining low computational complexity. The PSNR vs. FLOPs (G) compared with state-of-the-art methods are shown in Fig.~\ref{fig:glance}.

\section{Related Works}
\label{sec:relatedworks}
\subsubsection{Deep Image Deblurring}

Deep learning methods have achieved significant success in image deblurring \cite{Sun2015learning,Nah2017deep} as well as other low-level vision tasks such as image denoise \cite{Cheng2021nbnet,Zamir2020learning}, image deraining \cite{Jiang2020multi} and image super-resolution \cite{Mei2021image,Dong2016image,Zhang2018residual,Guo2020closed}. Sun \textit{et al.} \cite{Sun2015learning} propose to estimate the spatially-varying kernels of motion blur by a CNN. But, since the characteristics of blur are complex, the blur kernel estimation method is not practical in real scenarios. Later, DeepDeblur \cite{Nah2017deep} directly maps a blurry image to its sharp counterpart. Scale-recurrent network \cite{Tao2018scale} proposes an encoder-decoder structure to yield training feasibility. Adversarial training\cite{Kupyn2018deblurgan,Kupyn2019deblurgan,Zhang2020deblurring} and Recurrent Neural Networks\cite{Zhang2018dynamic,Park2020multi} also have been extensively studied. Most of these networks perform CNNs on the spatial domain to recover the sharp image.  
MAXIM \cite{Tu2022maxim} proposes MLP-based building blocks, which requires a big batch size for training. NAFNet \cite{Chen2022simple} designs computationally efficient networks from the baseline, which even achieves 33.69 dB PSNR on GoPro. Instead of designing a brand-new end-to-end image deblurring architecture, we reveal an intriguing phenomenon on frequency selection, and shed new light on improving the deblurring performance by incorporating faithful information about the blur pattern. 

Transformer/non-local has strong global context modeling ability and has shown its great promise in various computer vision tasks. Some transformer-based image restoration methods have been proposed, such as SwinIR \cite{Liang2021swinir}, Restormer \cite{Zamir2021restormer} and Uformer \cite{Wang2022uformer}. But the considerable computational complexity usually hampers their usage in efficient image restoration. We test the model of SwinIR \cite{Liang2021swinir} and Restormer (32.92 dB on GoPro) \cite{Zamir2021restormer} on GoPro dataset, which take 1.99s and 1.14s per image, respectively, even much slower than MPRNet, while our FMIMO-UNet (33.08 dB) takes 0.339s per image.

\subsubsection{End-to-end Deblur Model with ResBlock}

DeepDeblur \cite{Nah2017deep} designs a residual block (ResBlock) based on Conv-ReLU-Conv structure. Thereafter, ResBlock has become one fundamental block in image deblurring \cite{Tao2018scale,Zhang2019deep,Park2020multi,Cho2021rethinking}. Various efforts have been devoted to modifying the ResBlock, \textit{e.g.}, the content-aware processing module proposed by SAPHN \cite{Suin2020spatially}, the channel attention block proposed by MPRNet \cite{Zamir2021multi}, the HIN block proposed by HINet \cite{Chen2021hinet}, and the dilated conv block proposed by SDWNet \cite{Zou2021sdwnet}.

\subsubsection{Applications of Fourier Transform}

In recent years, some methods extract information from the frequency domain to fulfill different tasks \cite{Chi2019fast,Rippel2015spectral,Zhong2018joint,Yang2020fda,Rao2021global,Suvorov2022resolution}. FDA \cite{Yang2020fda} swaps the low-frequency spectrum between images to mitigate the influence caused by the images' style change for image segmentation. GFNet \cite{Rao2021global} learns long-term spatial dependencies in the frequency domain for image classification. LaMa \cite{Suvorov2022resolution} applies the structure of fast Fourier convolution \cite{Chi2020fast} to image inpainting. In image deblurring, SDWNet \cite{Zou2021sdwnet} introduces wavelet transform into deep networks. In this paper, we reveal an intriguing phenomenon of frequency selection for image deblurring.

\section{Method}
\label{sec:method}
\subsection{Empirical Findings of Frequency Selection via ReLU Operation}
We will describe the main observation insights of this paper in this section. The blurry image can be modeled as convolving the latent sharp image with the blur kernel. For simplicity, we elaborate the deviation in 1D scenario. The ``unknown'' sharp image is defined as $f(t)=\left\{\begin{matrix}
1 &  |t|<\tau \\
0 &  |t|>\tau\\
\end{matrix}\right.$ The blur kernel of a simple motion blur can be denoted as $g(t)=\delta(t)+\delta(t+\epsilon)$, where $\delta(\cdot)$ is the Direct delta function, and $\epsilon$ is a very small value.

\noindent\textbf{Remark 1}: Based on the above assumption, let's define a blurry image $b=g\otimes f$, where $\otimes$ means convolution, and we drop $t$ for simplicity. Taking the inverse Fourier transform (\emph{i.e.}, $\mathcal{F}^{-1}(\cdot)$) after ReLU (\emph{i.e.}, $\sigma(\cdot)$) on the blurry image in the frequency domain ($\mathcal{F}(\cdot)$ means Fourier transform) is written as $\hat{b}=\mathcal{F}^{-1}(\sigma(\mathcal{F}(b)))$, which can separate blur pattern component from other components. ReLU is applied to the real and imaginary parts respectively.

More details to support \textbf{Remark 1} are given in the supplementary material. One can easily decompose $\hat{b}$ and obtain a separate component from $\hat{b}$ as  $(\delta(t)+1/2(\delta(t-\epsilon)+\delta(t+\epsilon)))\otimes \mathcal{F}^{-1}(|\mathcal{F}(f(t))|)$. 

\begin{figure}[t]
\begin{center}
    \includegraphics[width=0.96\linewidth]{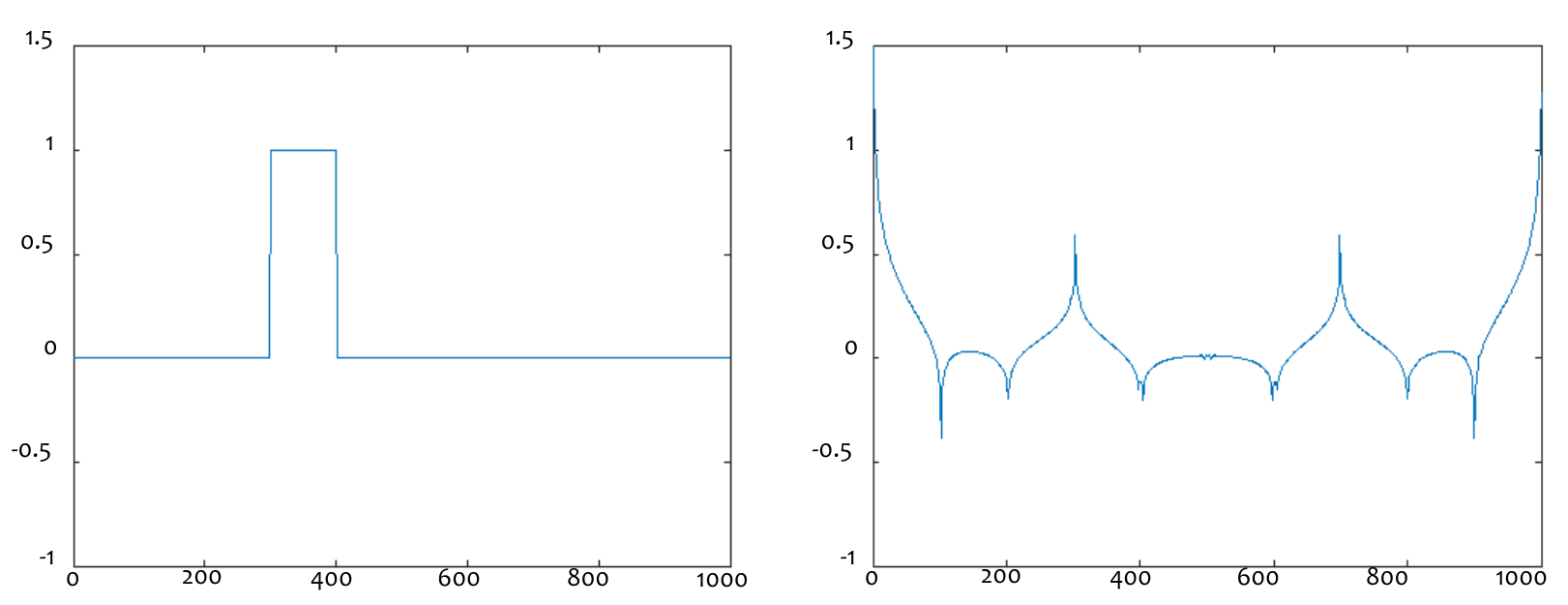}
\end{center}
\caption{Given a top-hat function $f$ on the left, its $\mathcal{F}^{-1}(|\mathcal{F}(f)|$ is plotted on the right, with obvious peaks, where $\mathcal{F}(\cdot)$ and $\mathcal{F}^{-1}(\cdot)$ mean discrete Fourier and inverse Fourier transform.} 
\label{fig:peak}
\end{figure}

\begin{figure*}[t]
\begin{center}
    \includegraphics[width=0.8\linewidth]{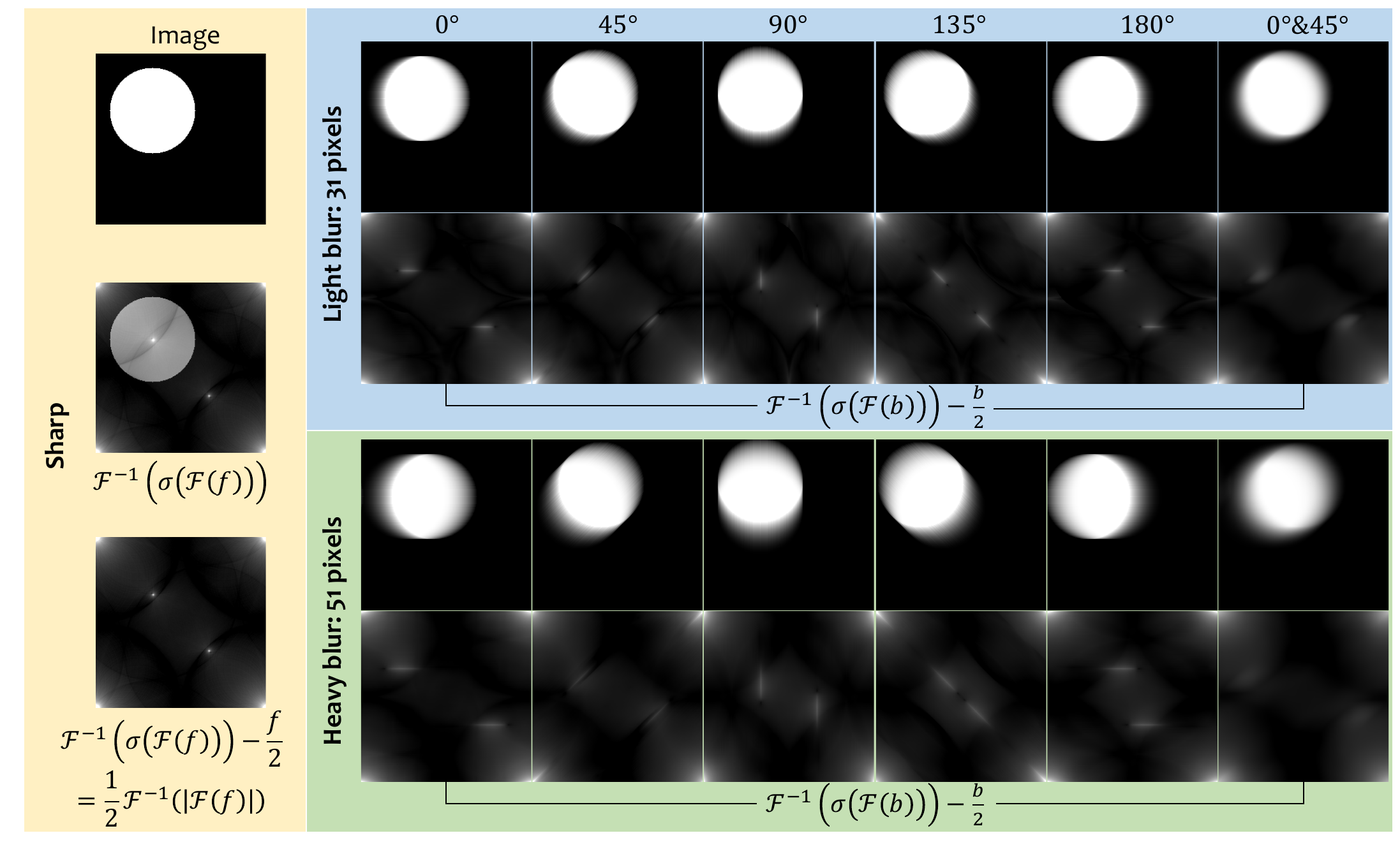}
\end{center}
\caption{We use a circle image as an example for sharp image $f$. The image is blurred ($b$) by linear kernels with motion blur of different directions $\{0^{\circ}, 45^{\circ}, 90^{\circ}, 135^{\circ}, 180^{\circ}\}$, different blur levels and a mixture of blur. Blur kernel-like images are calculated by the equations below the image. The middle image of the sharp image is visualized for reference. See more details in the supplementary material about equations $\mathcal{F}^{-1}(\sigma(\mathcal{F}(f)))-f/2$ and $\mathcal{F}^{-1}(\sigma(\mathcal{F}(b)))-b/2$ (best viewed  by zoom-in on screen).}
\label{fig:circle}
\end{figure*}

We observe that an absolute operation on Fourier transform of a top-hat function will produce peaks after inverse Fourier transform, as examples shown in Fig.~\ref{fig:peak}. We plot the discrete Fourier transform. Based on the phenomenon revealed in Fig.~\ref{fig:peak}, $\mathcal{F}^{-1}(\sigma(\mathcal{F}(g\otimes f)))$ will separate a component containing the convolution of $g$ and some peak values, when $f$ is a top-hat function.

Without loss of generality, we can extend \textbf{Remark 1} to more complicated scenarios. Fig.~\ref{fig:circle} shows examples of feature selection derived from a clean and blurry images caused by blur kernels with different blur direction, blur levels and mixtures of blur. As may be observed, taking inverse Fourier transform on frequency selection via ReLU acts as a sort of learning blur pattern directly from the blurry image. 

\begin{figure}[t]
\begin{center}
    \includegraphics[width=1\linewidth]{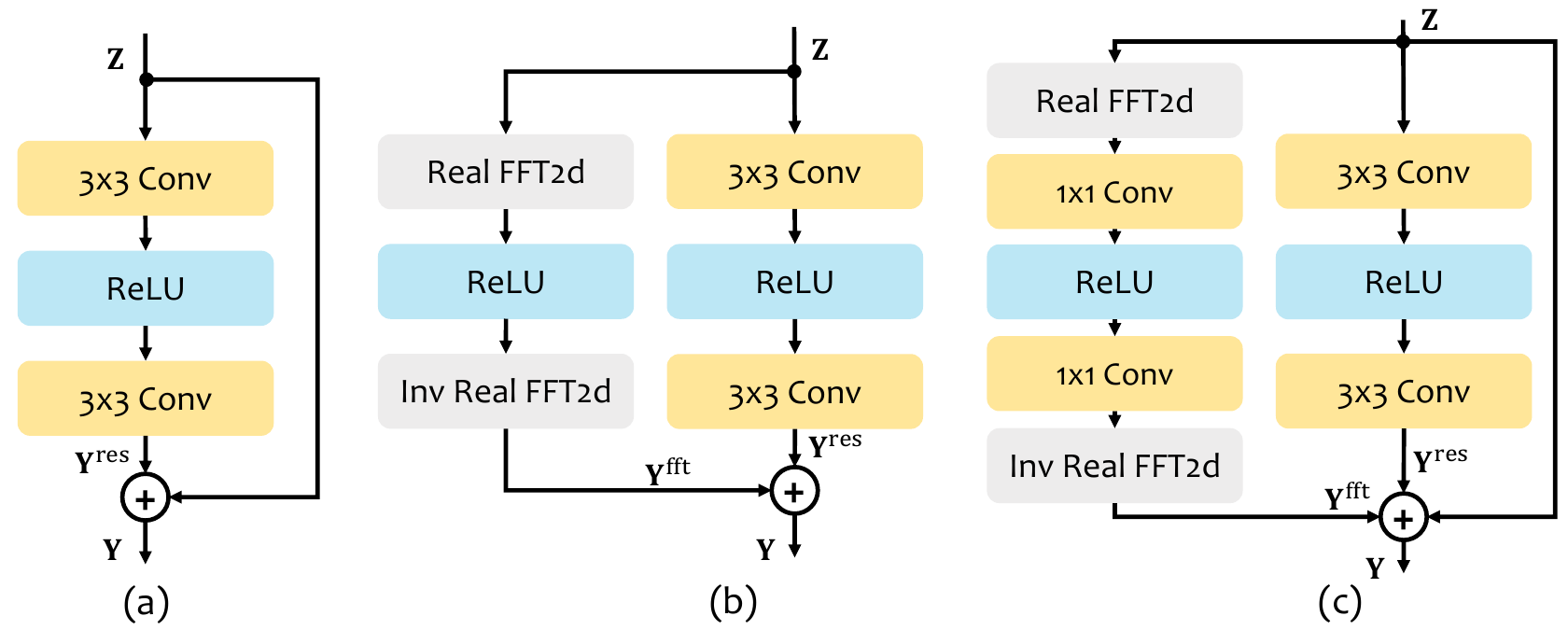}
\end{center}
\caption{(a) ResBlock. (b) Insert simple FFT-ReLU stream into ResBlock. (c) Proposed Res FFT-ReLU Block, where the bottom stream is termed as FFT-ReLU stream.} 
\label{fig:blockcompare}
\end{figure}

\subsection{Simple Fourier Transform with ReLU Stream}

\subsubsection{Basic Building Block for Image Deblurring}
The residual building block in image deblurring tasks is called ResBlock. Specifically, standard ResBlock learns pixel-level features, consisting of two $3\times 3$ convolutional layers and one ReLU layer in between, as shown in Fig.~\ref{fig:blockcompare} (a). Using inverse Fourier transform after selecting frequency from a blurry image $\mathcal{F}^{-1}(\sigma(\mathcal{F}(b)))$ generates blur patterns, implicitly showing the kernel pattern (see Eq. 3 in supplementary material and Fig.~\ref{fig:circle}, $\mathcal{F}^{-1}(\sigma(\mathcal{F}(b)))$ can be considered as a linear combination of a blur pattern image $\mathcal{F}^{-1}(|\mathcal{F}(b)|)$ and $b$). If such operations can be inserted into an end-to-end image deblurring network, the network will be able to learn both kernel-level and pixel-level information. Many methods can be applied to fuse kernel-level blur pattern features generated by $\mathcal{F}^{-1}(\sigma(\mathcal{F}(b)))$ and the pixel-level features. Instead of designing a complicated fusion block, we simply replace the \textit{identity} mapping by $\mathcal{F}^{-1}(\sigma(\mathcal{F}(\mathbf{Z})))$ for the sake of light computation, \emph{i.e.}, replace the \textit{identity} mapping in Fig.~\ref{fig:blockcompare}(a) by the left most stream, termed as simple-FFT-ReLU stream in Fig.~\ref{fig:blockcompare}(b). The position of simple-FFT-ReLU stream \emph{w.r.t.} the standard ResBlock (see Fig.~\ref{fig:simple-fft-relu}(a)) can be changed, and our experiments show that fusing kernel-level and pixel-level features either in parallel or sequentially can boost the results (see Table~\ref{tab:ablation-simplefftrelu-a}). 

\subsubsection{Analysis of ReLU on Frequency Domain from A New Perspective}
We analyze ReLU in frequency domain from another perspective. The phase and amplitude of a complex number $z=me^{j\beta}$ are $e^{j\beta}$ and $m\geq 0$ respectively. A Fourier transformed feature gives the phase and amplitude components. Applying ReLU (denoted as $\sigma(\cdot)$) on the frequency domain of a feature is defined as $\sigma(me^{j\beta})=m\sigma(e^{j\beta})$. This is considered as applying ReLU on the phase of the feature. As suggested by Pan \cite{Pan2019phase}, the phase of a blurry image plays an important role for deblurring, providing faithful information about the motion pattern. Applying ReLU indicates setting $T=0+j0$ as the selective threshold. What will happen if we change the threshold for frequency selection? As shown in Table~\ref{tab:ablation-simplefftrelu-b} in the experiment, setting different thresholds on ReLU, \emph{i.e.}, $m\sigma(e^{j\beta}-\frac{T}{m})+T$, gives different results, \emph{e.g.}, $T=100(1+j)$ leads to a better result compared with original ReLU. A proper threshold for selecting frequency is important for deblurring. But $T$ has to be set manually. To let the network modulate flexible thresholds for selecting frequencies, convolution can be further added after Fourier transform. For the sake of simplicity, let $a\cdot ke^{j\beta}+b$ denote the feature after $1\times$1 conv in frequency, where $a$ and $b$ are learnable complex values. $m\sigma(e^{j\beta}-\frac{T}{m})+T$ is then replaced by $m\sigma(a\cdot e^{j\beta}+\frac{b}{m})$ (ReLU after convolution).

\subsection{Res FFT-ReLU Block}
\label{sec:newblock}
From the above analysis, we propose a Residual Fast Fourier Transform with ReLU {Block} (Res FFT-ReLU Block) to replace the widely-used ResBlock. As shown in Fig.~\ref{fig:blockcompare}(c), different from Fig.~\ref{fig:blockcompare}(b), we keep the \emph{identity} mapping which assists network training. 

As shown in Fig.~\ref{fig:blockcompare}(c), besides a normal spatial residual stream, we simply add another stream based on a channel-wise FFT \cite{Brigham1967the}. DFT is widely used in modern signal processing algorithms, whose 1D version can be derived by $X[k]=\sum_{n=0}^{N-1}x[n]e^{-j\frac{2\pi}{N}kn}$,
where $x[n]$ is a sequence of $N$ complex numbers, $X[k]$ indicates the spectrum at the frequency $\omega_k=2\pi k/N$, and $j$ represent the imaginary unit. It is clear that the spectrum at any frequency has global information. Noted that the DFT of a real signal $x[n]$ is \textit{conjugate symmetric}, \textit{i.e.} $X[N-k]=\sum_{n=0}^{N-1}x[n]e^{-j\frac{2\pi}{N}(N-k)n}=X^*[k]$.
The same applies to 2D DFT, which performs sequential row and column 1D DFT on a 2D signal whose size is $M\times N$, \textit{i.e.}, $X[M-u,N-v]=X^*[u,v]$. Since the results of a real array's DFT has symmetric properties, the right half of the results can be derived from the left half. The FFT algorithms reduce the complexity and calculates the DFT in a more efficient way. Let $\mathbf{Z}\in\mathbb{R}^{H\times W\times C}$ be the input feature volume, where $H$, $W$, and $C$ indicate the height, width and channel of the feature volume. The left stream in Res FFT-ReLU Block, called FFT-ReLU stream (see Fig.~\ref{fig:blockcompare}(c)) is processed as follows:
\begin{itemize}
\item[(1)] computes 2D real FFT of $\mathbf{Z}$ and obtain $\tilde{\mathbf{Z}}=\mathcal{F}(\mathbf{Z})\in\mathbb{C}^{H\times W/2 \times C}$, where $\mathbb{C}$ means complex domain.
\item[(2)] uses two stacks of $1\times 1$ convolution layers (convolution operator $\otimes$) with a ReLU layer in between: $h(\tilde{\mathbf{Z}};\mathbf{\Theta}^{(1)},\mathbf{\Theta}^{(2)})=\text{ReLU}(\tilde{\mathbf{Z}}\otimes\mathbf{\Theta}^{(1)})\otimes\mathbf{\Theta}^{(2)}\in\mathbb{C}^{H\times W/2\times C}$, where $\mathbf{\Theta}^{(1)}, \mathbf{\Theta}^{(1)}\in\mathbb{C}^{C\times C}$ are parameters in complex values, and $h(\cdot;\mathbf{\Theta}^{(1)},\mathbf{\Theta}^{(2)})$ is the network block parameterized by $\mathbf{\Theta}^{(1)}$ and $\mathbf{\Theta}^{(2)}$. We apply ReLU to the real and imaginary parts respectively.
\item[(3)] applies inverse 2D real FFT to transform $h(\tilde{\mathbf{Z}};\mathbf{\Theta}^{(1)},\mathbf{\Theta}^{(2)})$ back to spatial domain: $\mathbf{Y}^\text{fft}=\mathcal{F}^{-1}\left(h(\tilde{\mathbf{Z}};\mathbf{\Theta}^{(1)},\mathbf{\Theta}^{(2)})\right)\in\mathbb{R}^{H\times W\times C}$.
\end{itemize}
Then the final output of Res FFT-ReLU Block is calculated via $\mathbf{Y}=\mathbf{Y}^\text{fft}+\mathbf{Y}^\text{res}+\mathbf{Z}$, where $\mathbf{Y}^\text{res}$ uses the same computation as that in the original ResBlock. The code of realizing $\Theta^{(1)}$ on $\mathbf{Z}$ is:
{\small\begin{verbatim}
weight_real=nn.Parameter(torch.Tensor(C, C))
weight_imag=nn.Parameter(torch.Tensor(C, C))
weight=torch.complex(weight_real,weight_imag)
Z_output=Z@weight #@ is matrix multiplication
\end{verbatim}
}

\subsection{Global Context Learning Ability}
\subsubsection{Convolution on Frequency Does Not Bring the Global Context Learning Ability}
\label{sec:discussion-conv}
Mathematically, given $x[n]$, which is a sequence of $N$ numbers. For $S\times 1$ convolution on the spatial domain, weights are denoted as $w[a]$, where $a\in\mathcal{N}_n(S)$, and $\mathcal{N}_n(S)$ indicates a $S\times 1$ neighborhood of $x[n]$. We have $y=\sum_{a}x[n-a]\cdot w[a]$. If we conduct $S\times 1$ convolution on the frequency domain, we have $y=\mathcal{F}^{-1}(\sum_aX[k-a]w[a])=\sum_aw[a]\cdot x[n]\cdot e^{j2\pi an/N}$, where $X[k]$ is from Res FFT-ReLU Block section. This indicates that linear convolution operation on frequency domain is only used to extract features, which cannot introduce global context learning ability to the network.

\subsubsection{ReLU on Frequency Brings the Global Context Learning Ability for the Network}
ReLU is a non-linear operation. The output feature on the $r$th spatial location after converting to the spatial domain is calculated by
$x[r]=\frac{1}{N}\sum_{k=0}^{N-1}\left(\sigma(\sum_{n=0}^{N-1}x[n]e^{-j2\pi kn/N})e^{j2\pi rk/N}\right)$.
Since ReLU is nonlinear, unlike convolution, it is not possible to simplify this equation by eliminating any $x[n]$. Thus, introducing ReLU in the frequency domain brings the global context learning ability for the network. We will show in Table~\ref{tab:ablation-simplefftrelu-a} that ReLU in frequency domain not always helps improve performance, whose location matters.

\section{Experiments}

\subsection{Experimental Setup}
\label{sec:dataset-detail}
\subsubsection{Dataset} Three datasets are mainly evaluated: GoPro \cite{Nah2017deep}, HIDE \cite{Shen2019human} and RealBlur \cite{Rim2020real} datasets. Since existing methods adopt different experimental settings, we summarize them and report two groups of results: \textit{({I})} train on 2,103 pairs of blurry and sharp images in GoPro dataset, and test on 1,111 image pairs in GoPro (follow \cite{Cho2021rethinking}), 2,025 image pairs in HIDE (follow \cite{Zamir2021multi}), 980 image pairs in RealBlur-R test set, and 980 image pairs in RealBlur-J test set (follow \cite{Zamir2021multi}), respectively; \textit{(II)} train on 3,758 image pairs in RealBlur-R, and test on 980 image pairs in RealBlur-R (follow \cite{Zamir2021multi}), and train on 3,758 image pairs in RealBlur-J, and test on 980 image pairs in RealBlur-J (follow \cite{Zamir2021multi}). Besides, we also show the effectiveness of Res FFT-ReLU Block on REDS dataset \cite{Nah2021ntire}, with 24,000 and 3,000 images for training and testing, respectively (follow \cite{Tu2022maxim}).

\subsubsection{Loss Function} We unify the loss function for all experiments. $\hat{\mathbf{S}}$, $\mathbf{S}$ and $\varepsilon$ denote the predicted sharp image, the groundtruth sharp image, and a constant value $10^{-3}$, respectively. Two kinds of loss functions are adopted: (1) {C}harbonnier loss \cite{Zamir2021multi}:
$\mathcal{L}_{c}=\sqrt{||\hat{\mathbf{S}}-{\mathbf{S}}||^2+\varepsilon^2}$, and (2) {F}requency {R}econstruction (FR) loss \cite{Cho2021rethinking,Tu2022maxim}: $\mathcal{L}_{fr}=||\mathcal{F}(\hat{\mathbf{S}})-\mathcal{F}(\mathbf{S})||_1$. Finally, the loss function is $\mathcal{L}=\mathcal{L}_{c}+\alpha_1\mathcal{L}_{fr}$, where $\alpha_1$ is a tradeoff-parameter and is empirically set to 0.01.

\subsubsection{Implementation Details} We adopt the training strategy used in MPRNet \cite{Zamir2021multi} unless otherwise specified. \textit{I.e.}, the network training hyperparameters (and the default values we use) are patch size (256$\times$ 256), batch size (16), training epoch (3,000), optimizer (Adam \cite{Kingma2015adam}), initial learning rate (2$\times$10$^{-4}$). The learning rate is steadily decreased to 1$\times$10$^{-6}$ using the cosine annealing strategy \cite{Loshchilov2017sgdr}. Following \cite{Zamir2021multi}, horizontal and vertical flips are randomly applied on patches for data augmentation. For testing, we adopt the same dataset slicing crop method as used in SDWNet \cite{Zou2021sdwnet}, where we utilize a step of 256 to perform 256$\times$256 size sliding window slicing, and compensate slicing on the edge part.

\subsubsection{Evaluation metric} The average performance of PSNR and SSIM over all testing sets are computed by using the official software released by \cite{Zamir2021multi}. We report number of parameters, FLOPs, and testing time per image (see supplementary material) on on a workstation with Intel Xeon Gold 6240C CPU, NVIDIA GeForce RTX 3090 GPU.

\subsection{Position Ablation on Simple-FFT-ReLU Stream}

\begin{figure}[t]
\begin{center}
    \includegraphics[width=0.9\linewidth]{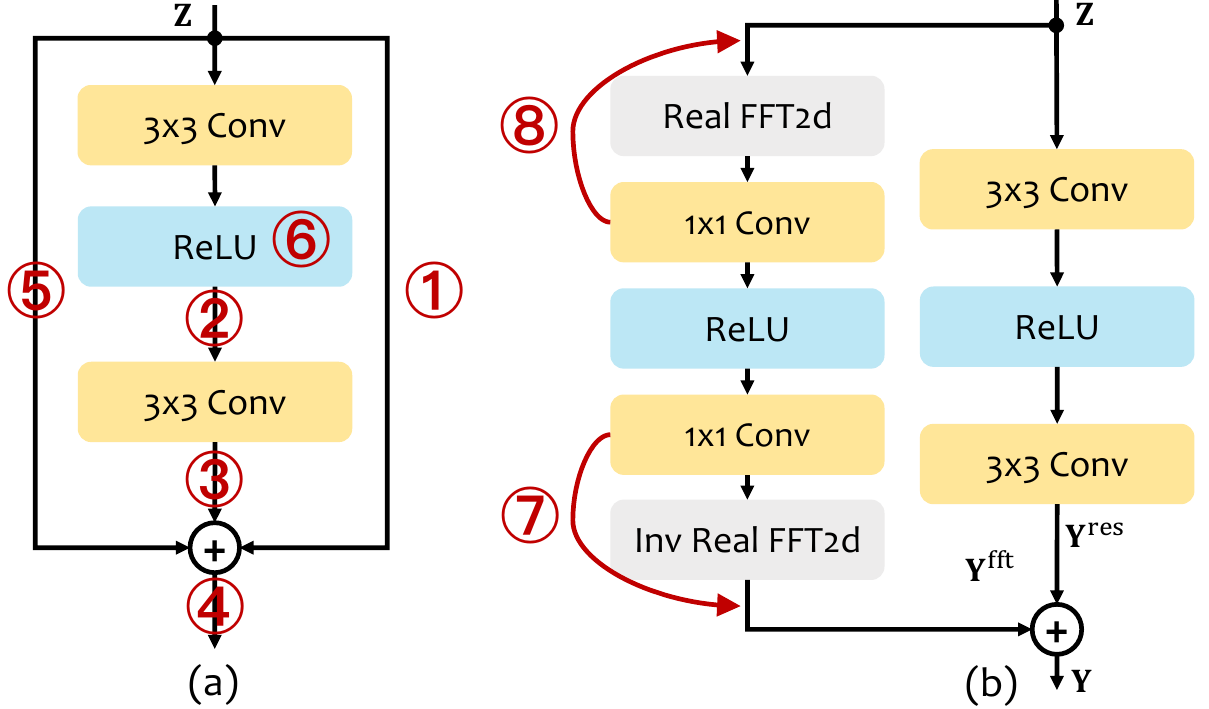}
\end{center}
\caption{Ablation on (a) the position of simple-FFT-ReLU, and (b) the position of 1$\times$1 convolution with simple-FFT-ReLU, where \ding{172} means we keep the \textit{identity} mapping, and \ding{177} indicates replacing ReLU with simple-FFT-ReLU. The others means the position where we insert simple-FFT-ReLU. \ding{178} and \ding{179} mean we put 1$\times$1 Conv before Real FFT2d or after inv Real FFT2d.} 
\label{fig:simple-fft-relu}
\end{figure}

We discuss how the position of simple-FFT-ReLU stream on the ResBlock changes the deblurring performance. Experiments are conducted on GoPro dataset (Group I setting). The backbone architecture we use is a simplified version of DeepDeblur \cite{Nah2017deep}, which contains 16 ResBlocks, and termed as RSNet in the paper. As shown in Fig.~\ref{fig:simple-fft-relu}(a), the position of simple-FFT-ReLU is changed \emph{w.r.t.} ResBlock. Results are reported in Table~\ref{tab:ablation-simplefftrelu-a}. If we insert simple-FFT-ReLU in postion \ding{177}, the performance drops, compared with RSNet (27.78 vs. 28.06). If we replace the \textit{identity} mapping by the simple-FFT-ReLU stream, PSNR increases compared with RSNet (29.51 vs. 28.06). This verifies our claim on the effectiveness of replacing \emph{identity} mapping by simple FFT-ReLU stream, as marked in \colorbox{gray!20}{gray} in Table~\ref{tab:ablation-simplefftrelu-a}. Based on this observation, we further change the threshold for filtering out frequencies. We conduct simple statistics and find that the real/imag values after FFT are mainly around $-10,000(1+j)$ to $+10,000(1+j)$. Thus, instead of setting $0+j0$ as the threshold, we try different thresholds, and also apply inverse ReLU (filter out frequency components larger than $0+j0$). Results are shown in Table~\ref{tab:ablation-simplefftrelu-b}. Inverse ReLU achieves similar results with ReLU (29.47 vs. 29.51). A positive threshold such as $+100(1+j)$ achieves better performance. Replacing ReLU by selecting high-/low- frequency components are not helpful at all, although all are non-linear operations.
We also conduct ablation study on the position of 1$\times$1 convolution with simple-FFT-ReLU (see Fig.~\ref{fig:simple-fft-relu} (b)). Results are shown in Table~\ref{tab:ablation-simplefftrelu-a}. 

\begin{table}[t]
\renewcommand\arraystretch{1}
\footnotesize
\centering
\caption{Ablation on GoPro dataset with RSNet \cite{Nah2017deep} for Fig.~\ref{fig:simple-fft-relu}(a) (upper area) and Fig.~\ref{fig:simple-fft-relu}(b) (bottom area). ``$\times$'' for \ding{176} means the left stream is totally removed in Fig.~\ref{fig:simple-fft-relu}(a). Result worse than RSNet is in \textit{italics}.}
\label{tab:ablation-simplefftrelu-a}
\resizebox{1\linewidth}{!}{
\begin{tabular}{cccccc|cc|cc}
\toprule[0.15em]
\ding{172} & \ding{173} & \ding{174} & \ding{175} & \ding{176} & \ding{177} & \ding{178} & \ding{179} & \textbf{PSNR} & \textbf{Params} (M)\\
\midrule
\checkmark & $\times$ & $\times$ & $\times$ & $\times$ & $\times$ & - & - & 28.06 & 0.30 \\
\checkmark & \checkmark & $\times$ & $\times$ & $\times$ & $\times$ & - & - & 29.17 & 0.30 \\
\checkmark & $\times$ & \checkmark & $\times$ & $\times$ & $\times$ & - & - & 29.77 & 0.30 \\
\checkmark & $\times$ & $\times$ & \checkmark & $\times$ & $\times$ & - & - & 29.73 & 0.30 \\
\checkmark & $\times$ & $\times$ & $\times$ & \checkmark & $\times$ & - & - & 29.08 & 0.30 \\
\rowcolor{gray!20}
$\times$ & $\times$ & $\times$ & $\times$ & \checkmark & $\times$ & - & - & 29.51 & 0.30 \\
\checkmark & $\times$ & $\times$ & $\times$ & $\times$ & \checkmark & - & - & \emph{27.78} & 0.30 \\
\midrule
- & - & - & - & - & - & $\times$ & $\times$ &  30.30 & 0.36\\
- & - & - & - & - & - & \checkmark & $\times$ & 30.17 & 0.35\\
- & - & - & - & - & - & \checkmark & \checkmark & 29.96 & 0.33 \\
\bottomrule[0.15em]
\end{tabular}
}
\end{table}

\begin{table}[t]
\renewcommand\arraystretch{1}
\footnotesize
\centering
\caption{Ablation on GoPro dataset with RSNet \cite{Nah2017deep}, for selecting frequency by different thresholds (upper), or replacing ReLU by high/low frequency selection (bottom) for the setting \colorbox{gray!20}{highlighted} in Table~\ref{tab:ablation-simplefftrelu-a}.}
\label{tab:ablation-simplefftrelu-b}
\begin{tabular}{c|c|c}
\toprule[0.15em]
\textbf{Thre} & \textbf{Freq} & ~\textbf{PSNR}~\\
\midrule
$-1000(1+j)$ & - &  28.71\\
$+1000(1+j)$ & - & 29.32  \\
$-100(1+j)$ & - & 28.78\\
$+100(1+j)$ & - & 29.60\\
inv. ReLU & - & 29.47\\
\midrule
- & HF (1/8) & \textit{27.81}\\
- & LF (1/8) & \textit{27.83}\\
\bottomrule[0.15em]
\end{tabular}
\end{table}

\begin{figure}[t]
\begin{center}
    \includegraphics[width=1\linewidth]{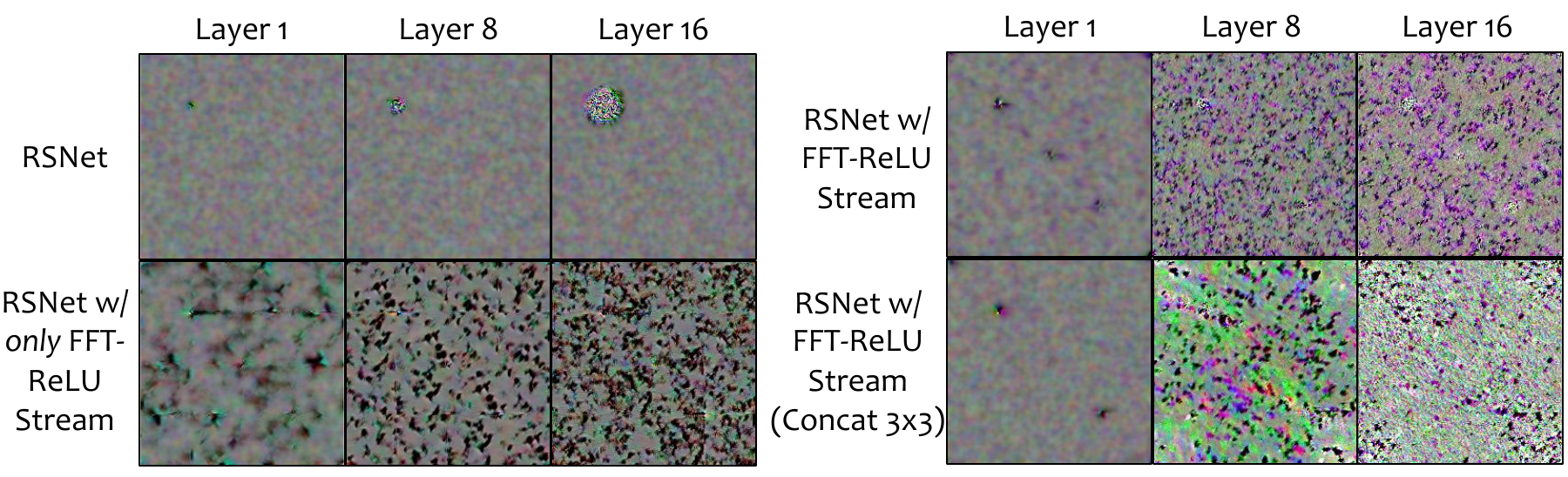}
\end{center}
\caption{Visualization of example features of three layers of different networks. In each layer, to maximize the neuron on location $[\frac{1}{4}H,\frac{1}{4}W]$, we show visualizations from one random gradient descent run for channel $C=16$.} 
\label{fig:neuron_vis}
\end{figure}

\subsection{Ablation Study on FFT-ReLU Stream}

\begin{table}[t]
\renewcommand\arraystretch{0.9}
\footnotesize
\centering
\caption{Ablation on FFT-ReLU stream on GoPro dataset with RSNet \cite{Nah2017deep}. ``Real'' means the weights in 1$\times$1 convolution are all real values, instead of the default complex values. ``Concat'' means after applying FFT, we concatenate the real part $\mathcal{R}(\mathcal{F}(\mathbf{Z}))$ and the imaginary part $\mathcal{I}(\mathcal{F}(\mathbf{Z}))$ along the channel dimension, so the weights of subsequent convolution operations are all real values. \colorbox{gray!20}{Gray} areas indicate RSNet and RSNet w/ FFT-ReLU stream. Results worse than RSNet are in \textit{italics}.}
\label{tab:ablation-fft-relu}
\resizebox{1\linewidth}{!}{
\begin{tabular}{cc|ccccc|cc}
\toprule[0.15em]
\multicolumn{2}{c}{\textbf{ResBlock}} & \multicolumn{5}{c}{\textbf{FFT-ReLU Stream}} & \textbf{PSNR} & \textbf{Params} \\
\cmidrule(lr){1-2} \cmidrule(lr){3-7}
$\mathbf{Z}$ & $\mathbf{Y}^\text{res}$ & FFT & Conv & ReLU & Conv & iFFT & & (M)\\
\midrule
\rowcolor{gray!20}
\checkmark & \checkmark & $\times$ & $\times$ & $\times$ & $\times$ & $\times$ & 28.06 & 0.30 \\
$\times$ & $\times$ & \checkmark & \checkmark & \checkmark & \checkmark & \checkmark & \textit{25.64} 
& 0.07\\
\checkmark & $\times$ & \checkmark & \checkmark & \checkmark & \checkmark & \checkmark & \textit{26.00} 
& 0.07\\
\checkmark & \checkmark & $\times$ & \checkmark & \checkmark & \checkmark & $\times$ & 28.16 & 0.33\\
\checkmark & \checkmark & \checkmark & $\times$ & \checkmark & $\times$ & \checkmark & 29.08 & 0.30\\
\checkmark & \checkmark & \checkmark & \checkmark & $\times$ & \checkmark & \checkmark & 28.59 & 0.36 \\
\checkmark & \checkmark & \checkmark & \checkmark & $\times$ & $\times$ & \checkmark & 28.60 & 0.33\\
\checkmark & \checkmark & \checkmark & Real & \checkmark & Real & \checkmark & 29.91 & 0.33\\
\checkmark & \checkmark & \checkmark & \checkmark & \checkmark & $\times$ & \checkmark & 30.06 & 0.33\\
\rowcolor{gray!20}
\checkmark & \checkmark & \checkmark & \checkmark & \checkmark & \checkmark & \checkmark & 30.30 & 0.36\\
\checkmark & \checkmark & \checkmark & \checkmark & GeLU & \checkmark & \checkmark & 30.42 & 0.36\\
\checkmark & \checkmark & \checkmark & Concat & \checkmark & Concat & \checkmark & 30.32 & 0.43\\
\checkmark & \checkmark & \checkmark & 3$\times$3 & \checkmark & 3$\times$3 & \checkmark & 31.15 & 1.48\\
\bottomrule[0.15em]
\end{tabular}
}
\end{table}

\subsubsection{Quantitative Results}
We conduct extensive ablation study on GoPro dataset (Group I setting) to have a thorough investigation on the proposed FFT-ReLU stream. We can summarize the following conclusions from Table~\ref{tab:ablation-fft-relu}. (1) Using RSNet with only our FFT-ReLU stream leads to obvious performance drop, compared with RSNet (25.64 vs. 28.06), but adding FFT-ReLU stream to RSNet leads to significant improvement (30.30 vs. 28.06). This indicates that our FFT-ReLU stream is an add on stream, which should be trained in conjunction with the pixel-level spatial domain CNN. We will illustrate the reason by visualizing neurons in the next subsection. (2) It is expected to see that w/o ReLU w/ two 1$\times$1 convolution layers, the performance does not change too much, compared with RSNet (28.59 vs. 28.06). (3) If we change the complex convolution into real convolution, the performance drops a little due to the decrease of parameter numbers (29.91 vs. 30.30). (4) Noted that we propose a complex convolution in frequency domain. An alternative is to first concatenates the real part $\mathcal{R}(\mathcal{F}(\mathbf{Z}))$ and the imaginary part $\mathcal{I}(\mathcal{F}(\mathbf{Z}))$ along the channel dimension to acquire $\tilde{\mathbf{Z}}=\mathcal{R}(\mathcal{F}(\mathbf{Z}))\odot_C \mathcal{I}(\mathcal{F}(\mathbf{Z}))\in\mathbb{R}^{H\times W/2\times 2C}$, where $\odot_C$ represents concatenation through the channel dimension. Then we can simply use two stacks of $1\times 1$ convolution layers (convolution operator $\otimes$) with a ReLU layer in between: $h(\tilde{\mathbf{Z}};\mathbf{\Theta}^{(1)}_\text{real},\mathbf{\Theta}^{(2)}_\text{real})=\text{ReLU}(\tilde{\mathbf{Z}}*\mathbf{\Theta}^{(1)}_\text{real})\otimes\mathbf{\Theta}^{(2)}_\text{real}\in\mathbb{R}^{H\times W/2\times 2C}$. The inverse 2D real FFT can be applied afterwards. We term this alternative as ``Concat''. Using Concat, the weights in convolution are all real values, and the performance does not change compared with our complex convolution (30.32 vs. 30.30), but the number of parameters increases. (5) We further replace 1$\times$1 convolution by 3$\times$3 convolution, which can be easily implemented in ``Concat'' setting. A further improvement is oberved (31.15 vs. 30.32). But, Concat 3$\times$3 increases the model parameters by 4 times, compared with Concat 1$\times$1.

\subsubsection{Visualizations of Neurons}
We visualize neurons to better understand how our FFT-ReLU stream works. In \cite{Yosinski2015understanding}, the learned features computed by individual neurons at any layer of the network can be visualized by generating the input image such that the corresponding neurons activation value is maximized. Similarly, given a feature volume $\mathbf{Z}_i\in\mathbb{R}^{H\times W\times C}$ in Layer $i$, where $H$, $W$ and $C$ mean the height, width and the channel, we show visualizations of neurons on spatial location of $[\frac{1}{4}H, \frac{1}{4}W]$ from one random gradient descent run for $C=16$ in Layer 1, 8, 16, using a public repository\footnote{https://github.com/utkuozbulak/pytorch-cnn-visualizations}. As shown in Fig.~\ref{fig:neuron_vis}, to maximize the neuron on $[\frac{1}{4}H, \frac{1}{4}W]$, RSNet only gathers information from its local neighborhood. RSNet w/ \textit{only} FFT-ReLU stream can learn global context, but using only FFT-ReLU stream lacks pixel-level localization ability, \textit{i.e.}, there is not a clear activation region on $[\frac{1}{4}H, \frac{1}{4}W]$. This explains the reason why it can only obtain $26.00$ dB in PSNR in Table~\ref{tab:ablation-fft-relu}.  RSNet w/ FFT-ReLU stream can learn both kernel-level and pixel-level representations. Another interesting observation is that though FFT-ReLU stream has the ability to learn global context, in the lower layers, \textit{e.g.}, Layer 1, local information is more important, like Transformer \cite{Raghu2021do}. Other details are provided in the supplementary material.

\subsection{{Evaluation of Res FFT-ReLU Block}}

\begin{table}[t]
\renewcommand\arraystretch{0.9}
\footnotesize
\centering
\caption{Evaluation of Res FFT-ReLU Block (Group I setting) in PSNR. FFT means the ResBlock is replaced by Res FFT-ReLU Block. $\times$ means the original architecture. All models are trained by ourselves for fair comparison.}
\label{tab:fft}
\resizebox{1\linewidth}{!}{
\begin{tabular}{lc|ll|cc}
\toprule[0.15em]
\textbf{Model} & \textbf{FFT} & \textbf{GoPro} & \textbf{HIDE} & \textbf{Params} \scriptsize{(M)} & \textbf{FLOPs} \scriptsize{(G)} \\
\midrule[0.09em]
Deep Deblur & $\times$ & 31.15 & 29.17 & 11.71 & 336.03\\
& \checkmark & 32.37{\color{cyan}\textbf{\scriptsize{$\uparrow$}1.22}} & 30.86{\color{cyan}\textbf{\scriptsize{$\uparrow$}1.69}} & 12.65 & 363.19\\
\arrayrulecolor{black!30}\midrule[0.02em]
{RSNet} & $\times$ & 28.06 & 26.01 & 0.30 & 19.48\\
 & \checkmark & 30.30{\color{cyan}\textbf{\scriptsize{$\uparrow$}2.24}} & 28.87{\color{cyan}\textbf{\scriptsize{$\uparrow$}2.86}} & 0.36 & 23.84\\
\arrayrulecolor{black!30}\midrule[0.02em]
{U-Net} & $\times$ & 29.20 & 27.15 & 0.62 & 12.18\\
 & \checkmark & 30.39{\color{cyan}\textbf{\scriptsize{$\uparrow$}1.19}} & 28.71{\color{cyan}\textbf{\scriptsize{$\uparrow$}1.56}} & 0.76 & 14.86\\
\arrayrulecolor{black!30}\midrule[0.02em]
MPRNet & $\times$ & 31.09 & 29.66 & 2.14 & 80.70 \\
-small& \checkmark & 32.50{\color{cyan}\textbf{\scriptsize{$\uparrow$}1.41}} & 30.82{\color{cyan}\textbf{\scriptsize{$\uparrow$}1.16}} & 2.43 & 94.24 \\
\arrayrulecolor{black!30}\midrule[0.02em]
MIMO-UNet & $\times$ & 31.90 & 29.62 & 6.80 & 67.17\\
 & \checkmark & 32.71{\color{cyan}\textbf{\scriptsize{$\uparrow$}0.81}} & 30.85{\color{cyan}\textbf{\scriptsize{$\uparrow$}1.23}} & 8.17 & 80.21\\
\arrayrulecolor{black!30}\midrule[0.02em]
NAFNet32 & $\times$ & 32.95 & 30.60 & 17.1 & 16.00\\
 & \checkmark & 33.12{\color{cyan}\textbf{\scriptsize{$\uparrow$}0.17}} & 30.76{\color{cyan}\textbf{\scriptsize{$\uparrow$}0.16}} & 17.8 & 23.10\\
\arrayrulecolor{black}\bottomrule[0.15em]
\end{tabular}
}
\end{table}

\subsubsection{Quantitative Results}

Res FFT-ReLU Block is plug-and-play. We plug it into various architectures: DeepDeblur \cite{Nah2017deep}; U-Net (one backbone network used in MPRNet \cite{Zamir2021multi}); MPRNet-small \cite{Zamir2021multi}, whose number of channels is three times smaller than original MPRNet due to limited computation resource; MIMO-UNet \cite{Cho2021rethinking}; NAFNet \cite{Chen2022simple}. NAFNet uses a different ResBlock, and we show the building block of NAFNet w/ Res FFT-ReLU Block in the supplementary material. 
PSNRs on GoPro and HIDE datasets in Group I setting are summarized in Table~\ref{tab:fft}. Replacing ResBlock by our Res FFT-ReLU Block leads to remarkable performance gains in various architectures. Besides, RSNet and RSNet w/ Res FFT-ReLU Block on \textbf{REDS} dataset are tested. Results are 26.78 and 27.79{\color{cyan}\textbf{\scriptsize{$\uparrow$}1.01}} dB respectively.

\subsection{Evaluation of FMIMO-UNet and FNAFNet}

\begin{table}[!t]
\begin{center}
\caption{Comparison on GoPro, HIDE and RealBlur datasets (Group I and II settings). Group II results are with {\color{red}$\mathparagraph$}.}
\label{tab:deblurring}
\renewcommand\arraystretch{0.8}
\setlength{\tabcolsep}{1.9pt}
\resizebox{1\linewidth}{!}{
\begin{tabular}{l c | c | c | c }
\toprule[0.15em]
 & \textbf{GoPro} & \textbf{HIDE} & \textbf{RealBlur-R} & \textbf{\textbf{RealBlur-J}} \\
 \textbf{Method} & PSNR~\colorbox{color4}{SSIM} & PSNR~\colorbox{color4}{SSIM} & PSNR~\colorbox{color4}{SSIM} & PSNR~\colorbox{color4}{SSIM}\\
\midrule[0.15em]
Xu \emph{et al.} (CVPR'13) & 21.00 \colorbox{color4}{0.741} & -  &   34.46 \colorbox{color4}{0.937} &  27.14 \colorbox{color4}{0.830} \\
Hu \emph{et al.} (CVPR'14) & - & - & 33.67 \colorbox{color4}{0.916} & 26.41 \colorbox{color4}{0.803}\\
Pan \emph{et al.} (CVPR'16) & - & - & 34.01~\colorbox{color4}{0.916} & 27.22~\colorbox{color4}{0.790}\\
Nah \emph{et al.} (CVPR'17)  & 29.08 \colorbox{color4}{0.914} & 25.73 \colorbox{color4}{0.874}  &  32.51 \colorbox{color4}{0.841}  &  27.87 \colorbox{color4}{0.827} \\
SRN~(CVPR'18)  & 30.26 \colorbox{color4}{0.934} & 28.36 \colorbox{color4}{0.915} & 35.66 \colorbox{color4}{0.947} &  28.56  \colorbox{color4}{0.867} \\
Zhang \emph{et al.} (CVPR'18) & 29.19 \colorbox{color4}{0.931} & - &  35.48 \colorbox{color4}{0.947}  &  27.80 \colorbox{color4}{0.847}\\
DeblurGAN (CVPR'18) & 28.70 \colorbox{color4}{0.858} & 24.51 \colorbox{color4}{0.871} & 33.79 \colorbox{color4}{0.903}  &  27.97 \colorbox{color4}{0.834} \\
\small{DeblurGAN-v2 (ICCV'19)}    & 29.55 \colorbox{color4}{0.934} & 26.61 \colorbox{color4}{0.875} & 35.26 \colorbox{color4}{0.944}  &  28.70 \colorbox{color4}{0.866} \\
Gao \emph{et al.} (CVPR'19) & 30.90 \colorbox{color4}{0.935} &  29.11 \colorbox{color4}{0.913}  & -      & - \\
DMPHN (CVPR'19) & 31.20 \colorbox{color4}{0.940}  & 29.09 \colorbox{color4}{0.924} &  35.70 \colorbox{color4}{0.948} & 28.42 \colorbox{color4}{0.860} \\
RADN (AAAI'20) & 31.76 \colorbox{color4}{0.953} & - & - & -\\
Suin \emph{et al.} (CVPR'20) & 31.85 \colorbox{color4}{0.948} & 29.98 \colorbox{color4}{0.930} & -       & - \\
SVDN (CVPR'20) & 29.81 \colorbox{color4}{0.937} & - & - & -\\
DBGAN (CVPR'20) & 31.10 \colorbox{color4}{0.942} & 28.94 \colorbox{color4}{0.915}  & 33.78 \colorbox{color4}{0.909}     & 24.93 \colorbox{color4}{0.745} \\
MT-RNN (ECCV'20) & 31.15 \colorbox{color4}{0.945} & 29.15 \colorbox{color4}{0.918}   & 35.79 \colorbox{color4}{0.951}     & 28.44 \colorbox{color4}{0.862}\\
MPRNet (CVPR'21) & 32.66 \colorbox{color4}{0.959} & {30.96} \colorbox{color4}{0.939} & {35.99} \colorbox{color4}{0.952} & {28.70} \colorbox{color4}{0.873}\\
MIMO-UNet (ICCV'21) & 31.73 \colorbox{color4}{0.951} & 29.28 \colorbox{color4}{0.921} & 35.47 \colorbox{color4}{0.946}& 27.76 \colorbox{color4}{0.836}\\
MIMO-UNet+~(ICCV'21) & 32.45 \colorbox{color4}{0.957} & 29.99 \colorbox{color4}{0.930} & 35.54 \colorbox{color4}{0.947} & 27.63 \colorbox{color4}{0.837}\\
\small{MIMO-UNet++ (ICCV'21)} & 32.68 \colorbox{color4}{0.959} & - & - & -\\
SPAIR~(ICCV'21) & 32.06 \colorbox{color4}{0.953} & 30.29 \colorbox{color4}{0.931} & - & {28.81} \colorbox{color4}{{0.875}}\\
HINet (CVPRW'21) & {32.71} \colorbox{color4}{0.959} & - & - & -\\
Restormer (CVPR'22) & {32.92} \colorbox{color4}{{0.961}} & {31.22} \colorbox{color4}{{0.942}} & {36.19} \colorbox{color4}{{0.957}} & {28.96} \colorbox{color4}{{0.879}}\\
Uformer (CVPR'22) & 33.06 \colorbox{color4}{0.967} & 30.90 \colorbox{color4}{0.953} & 36.19 \colorbox{color4}{0.956} & 29.09 \colorbox{color4}{0.886}\\
NAFNet32 (ECCV'22) & 32.85 \colorbox{color4}{0.959} & 30.60 \colorbox{color4}{0.936} & 35.97 \colorbox{color4}{0.951} & 28.75 \colorbox{color4}{0.875}\\
NAFNet64 (ECCV'22) & 33.69 \colorbox{color4}{0.967} &31.32  \colorbox{color4}{0.943} & - & - \\
\arrayrulecolor{black!30}\midrule
FMIMO-UNet-small & 32.50 \colorbox{color4}{0.958} & 30.65 \colorbox{color4}{0.937} & 35.92 \colorbox{color4}{0.952} & 28.58 \colorbox{color4}{0.864}\\
FMIMO-UNet & 33.08 \colorbox{color4}{0.962} & 31.19 \colorbox{color4}{0.943} & 35.96 \colorbox{color4}{0.953} & 28.72 \colorbox{color4}{0.871}\\
FMIMO-UNet+ & 33.52 \colorbox{color4}{0.965} & \textbf{31.66} \colorbox{color4}{0.946} & 36.11 \colorbox{color4}{0.955} & 28.88 \colorbox{color4}{0.880}\\
FNAFNet32 & 33.12 \colorbox{color4}{0.962} & 30.76 \colorbox{color4}{0.938} & 36.07 \colorbox{color4}{0.955} & 28.78 \colorbox{color4}{0.879}\\
FNAFNet64 & \textbf{33.85} \colorbox{color4}{0.967} & 31.12 \colorbox{color4}{0.944} & - & - \\
\arrayrulecolor{black}\bottomrule[0.1em]
\bottomrule[0.1em]
\small{DeblurGAN-v2 {\color{red}$\mathparagraph$} (ICCV'19)} & - & - & 36.44 \colorbox{color4}{0.935} & 29.69 \colorbox{color4}{0.870}\\
SRN {\color{red}$\mathparagraph$} (CVPR'18) & - & - & 38.65 \colorbox{color4}{0.965} & 31.38 \colorbox{color4}{0.909}\\
MPRNet {\color{red}$\mathparagraph$} (CVPR'21) & - & - & {39.31} \colorbox{color4}{0.972} & 31.76 \colorbox{color4}{0.922}\\
\arrayrulecolor{black!30}\midrule
FMIMO-UNet {\color{red}$\mathparagraph$} & - & - & 40.01 \colorbox{color4}{0.972} & \textbf{32.65} \colorbox{color4}{0.931}\\
FMIMO-UNet+ {\color{red}$\mathparagraph$} & - & - & \textbf{40.01} \colorbox{color4}{0.973} & 32.63 \colorbox{color4}{0.933} \\
\arrayrulecolor{black}\bottomrule[0.15em]
\end{tabular}}
\end{center}
\end{table}

MIMO-UNet and NAFNet are most recent ResBlock-based models with remarkable speed advantages among existing networks. We design FMIMO-UNet based on MIMO-UNet and MIMO-UNet+, acquiring FMIMO-UNet and FMIMO-UNet+. Besides, we reduce the number of Res FFT-ReLU Block to 4 for each encoder block and decoder block, and obtain an extra model termed as FMIMO-UNet-small. We design FNAFNet based on NAFNet. FMIMO-UNets and FNAFNet are compared with other state-of-the-arts in Table~\ref{tab:deblurring}. Note that due to the limited computation resource, Res FFT-ReLU block is only added to a few encoder/decoder blocks. We set the batch size to 32 for both FNAFNet32 and FNAFNet64. See detailed analysis, computational cost comparisons and visualizations in the supplementary material.

\section{Conclusion}
In this paper, we reveal an intriguing phenomenon that frequency selection can provide faithful information about the blur pattern. With this viewpoint, we propose a plug-and-play block called Res FFT-ReLU Block based on FFT-ReLU stream. Res FFT-ReLU Block allows the image-wide receptive field which is able to capture the long-term interaction. Plugging Res FFT-ReLU Block into MIMO-UNet and NAFNet achieves remarkable superior performance compared with state-of-the-arts, on three well-known public image deblurring datasets.

\section{Acknowledgments}
This work was supported by the National Natural Science Foundation of China (Grant No. 62101191, 61975056), Shanghai Natural Science Foundation (Grant No. 21ZR1420800), and the Science and Technology Commission of Shanghai Municipality (Grant No. 20440713100, 22DZ2229004).

\bibliography{aaai23}

\clearpage

\appendix

\section{Appendix}

\subsection{Analysis of Remark 1}

In this section, we provide the \emph{Analysis} for Remark 1. For simplicity, we elaborate the deviation in 1D scenario. The ``unknown'' sharp image is defined as $f(t)=\left\{\begin{matrix}
1 &  |t|<\tau \\
0 &  |t|>\tau\\
\end{matrix}\right.$ The blur kernel of a simple motion blur can be denoted as $g(t)=\delta(t)+\delta(t+\epsilon)$, where $\delta(\cdot)$ is the Direct delta function, and $\epsilon$ is a very small value.

\emph{Analysis.} Taking the inverse Fourier transform (denoted as $\mathcal{F}^{-1}(\cdot)$) after ReLU (denoted as $\sigma(\cdot)$, applied on real part and imaginary part separately) on the blurry image (denoted as $b$) in the frequency domain (Fourier transform is denoted as $\mathcal{F}(\cdot)$) can be formulated as:
\begin{eqnarray}
&&\hat{b}=\mathcal{F}^{-1}(\sigma\left(\mathcal{F}(b)\right))\\ 
&&=\mathcal{F}^{-1}\left(\frac{\mathcal{F}(b)+\left|\mathcal{F}(b)\right|}{2}\right)\\
&&=\frac{b}{2}+\frac{1}{2}\mathcal{F}^{-1}(|\mathcal{F}(b)|)\\
&&=\frac{b}{2}+\frac{1}{2}\mathcal{F}^{-1}(\left|\mathcal{F}(g(t)\otimes f(t))\right|)\\ 
&&=\frac{b}{2}+ \frac{1}{2}\mathcal{F}^{-1}\left(\left|\mathcal{F}(g(t))\cdot\mathcal{F}(f(t))\right|\right)\\ 
&&=\frac{b}{2}+\frac{1}{2}\mathcal{F}^{-1}(\left|\big(1+\cos(\omega \epsilon )+j\sin(\omega \epsilon)\big)\cdot\mathcal{F}(f(t))\right|)\\\nonumber 
&&=\frac{b}{2}+\frac{1}{2}\mathcal{F}^{-1}\left((1+\cos(\omega \epsilon))|\mathcal{F}(f(t))|\right)\\
&&~~~+\frac{1}{2}\mathcal{F}^{-1}(\left|j\sin(\omega\epsilon)\mathcal{F}(f(t))\right|)\\\nonumber
&&\footnotesize{=\frac{1}{2}\left(\delta(t)+\frac{1}{2}\left[\delta(t-\epsilon)+\delta(t+\epsilon)\right]\right)\otimes \mathcal{F}^{-1}(|\mathcal{F}(f(t))|)}\\
&&+\frac{1}{2}\mathcal{F}^{-1}(\left|j\sin(\omega\epsilon)\mathcal{F}(f(t))\right|)+\frac{b}{2}.
\end{eqnarray}
Note that $|\cdot|$ denotes conducting \emph{absolute} operation on real part and imaginary part separately. According to Eq. 8, one can simply extract component $\left(\delta(t)+\frac{1}{2}\left[\delta(t-\epsilon)+\delta(t+\epsilon)\right]\right)\otimes \mathcal{F}^{-1}(|\mathcal{F}(f(t))|)$ from $\hat{b}$, which contains blur pattern. As shown in Fig. 2 in the paper, $\mathcal{F}^{-1}(|\mathcal{F}(f(t))|)$ produces peaks, which are caused by the $|\cdot|$ operation. Thus, $\hat{b}$ contains a blur pattern component: the convolution of the blur kernel $g(t)$ and some decomposed peak values. This component implicitly shows the kernel pattern.

\subsection{Supplementary for Experiments}

\subsubsection{Network Structure}

\paragraph{The Structure of RSNet}

\begin{figure}[h]
\begin{center}
    \includegraphics[width=1\linewidth]{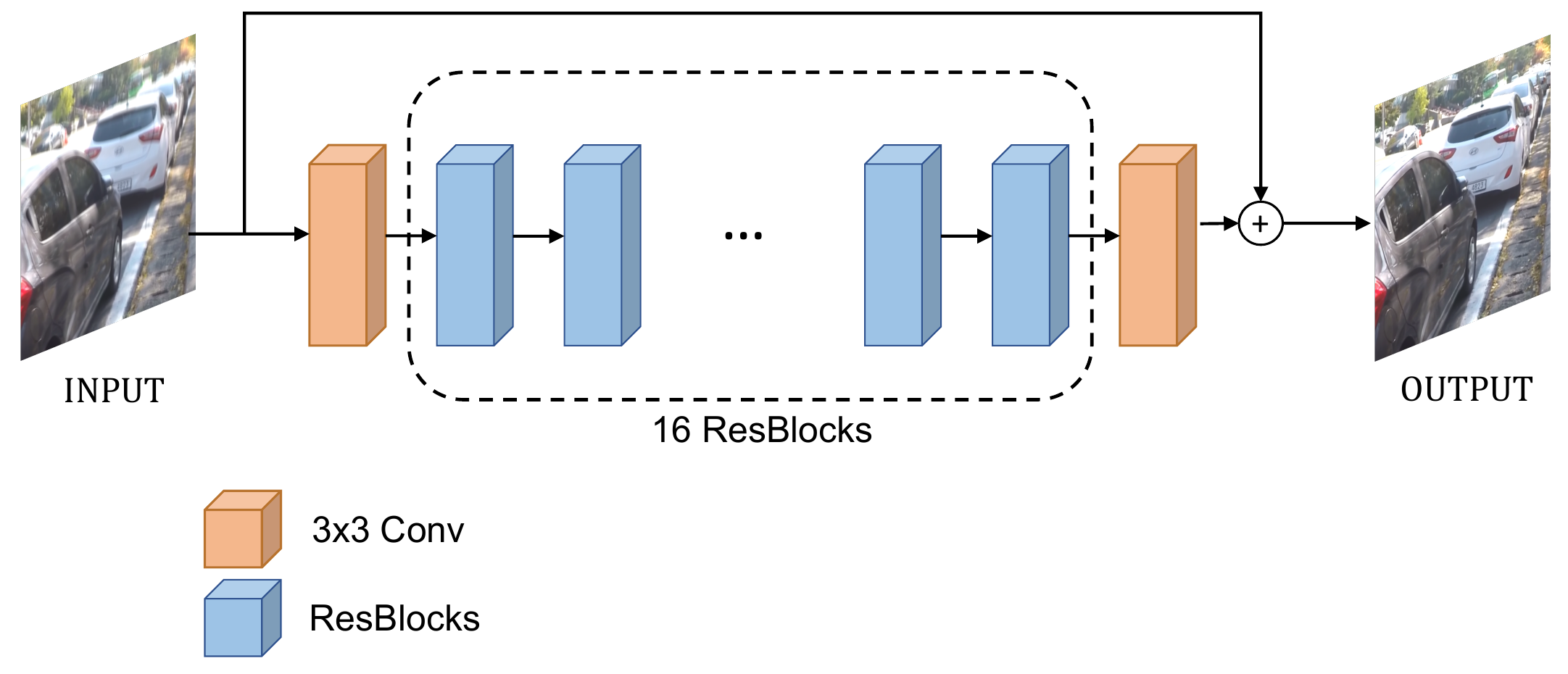}
\end{center}
\caption{The structure of RSNet.} 
\label{fig:rsnet}
\end{figure}

The backbone architecture we use in our ablation study is a simplified version of DeepDeblur \cite{Nah2017deep}, which contains 16 ResBlocks. We term it as RSNet, whose structure is illustrated in Fig.~\ref{fig:rsnet}.

\paragraph{Building Block of NAFNet w/ Res FFT-ReLU}

\begin{figure}[t]
\begin{center}
    \includegraphics[width=0.55\linewidth]{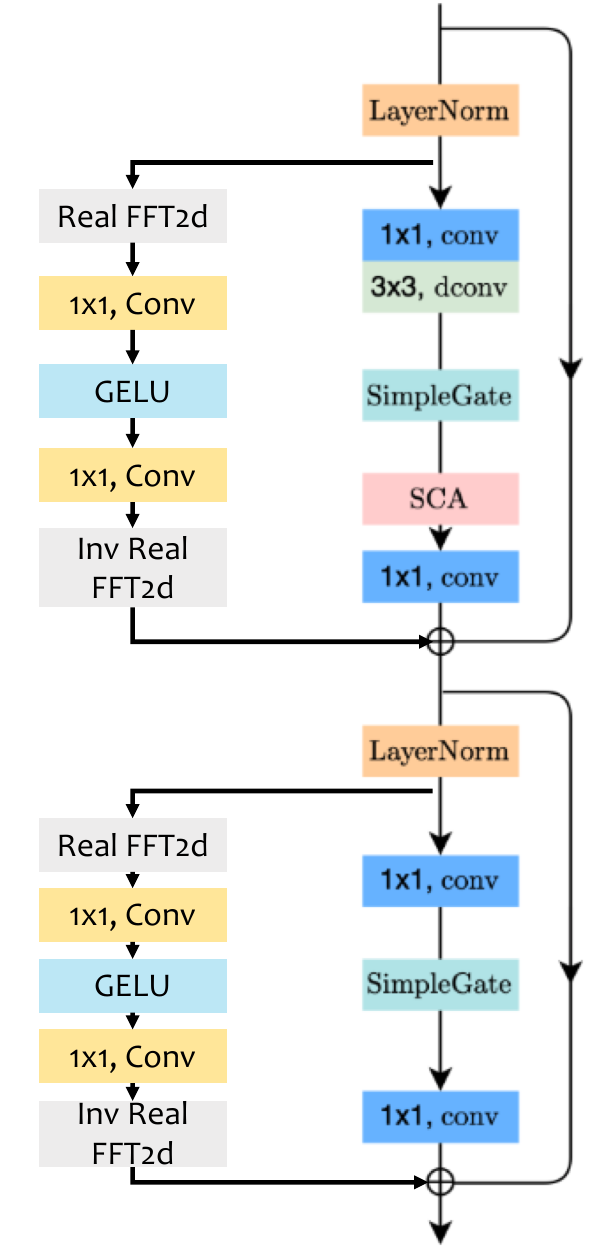}
\end{center}
\caption{Building block of NAFNet w/ Res FFT-ReLU.} 
\label{fig:nafnetblock}
\end{figure}

NAFNet \cite{Chen2022simple} uses a different ResBlock compared with Fig. 4(a) in the paper. Thus, we add the proposed FFT-ReLU stream in a slightly different way. As shown in Fig.~\ref{fig:nafnetblock}, the proposed Res FFT-ReLU block of FNAFNet consists of two FFT-ReLU streams.

In NAFNet, the computation is mainly conducted in the $5$th blocks. Noted that because of our limited computation resource, for FNAFNet32, we add Res FFT-ReLU block to the first three encoder blocks and their skip connected decoder blocks. For FNAFNet64, we only add Res FFT-ReLU block to the first encoder block and its skip connected decoder block. 

\begin{table*}[t]
\renewcommand\arraystretch{0.9}
\footnotesize
\centering
\caption{Ablation on FFT-ReLU stream on GoPro dataset \cite{Nah2017deep} with RSNet \cite{Nah2017deep} (16 ResBlocks). Let's assume the height of a feature channel is $H$. ``HF (1/4)'' or ``LF (1/4)'' indicates after FFT, we only keep the high-frequency component which is outside or low-frequency component which is inside the circle with the radius of $\frac{1}{4}H$, respectively. ``Layer 1 $\sim$ Layer 8'' denotes we only add FFT-ReLU stream on the first 8 ResBlocks. \colorbox{gray!20}{Gray} areas indicate conventional RSNet and RSNet w/ FFT-ReLU stream, as shown in Fig.~\ref{fig:highlowfreq}. Results worse than conventional RSNet are in \textit{italics}.}
\label{tab:frequencymultilayer}
\resizebox{0.7\linewidth}{!}{
\begin{tabular}{cc|ccccc|cc}
\toprule[0.15em]
\multicolumn{2}{c|}{ResBlock} & \multicolumn{5}{c|}{FFT-ReLU Stream} \\
\cmidrule(lr){1-2} \cmidrule(lr){3-7}
~~$\mathbf{Z}$~~ & ~~$\mathbf{Y}^\text{res}$~~ & ~~FFT~~ & ~~Conv~~ & ~~ReLU~~ & ~~Conv~~ & ~~iFFT~~ & ~~PSNR~~ & ~~Params (M)~~\\
\midrule
\rowcolor{gray!20}
\checkmark & \checkmark & $\times$ & $\times$ & $\times$ & $\times$ & $\times$ & 28.06 & 0.30 \\
\rowcolor{gray!20}
\checkmark & \checkmark & \checkmark & \checkmark & \checkmark & \checkmark & \checkmark & 30.30 & 0.36\\
\midrule
\checkmark & \checkmark & HF (1/4) & \checkmark & \checkmark & \checkmark & \checkmark & 28.64 & 0.36\\
\checkmark & \checkmark & LF (1/4) & \checkmark & \checkmark & \checkmark & \checkmark & 30.44 & 0.36\\
\checkmark & \checkmark & HF (1/8) & \checkmark & \checkmark & \checkmark & \checkmark & 29.03 & 0.36\\
\checkmark & \checkmark & LF (1/8) & \checkmark & \checkmark & \checkmark & \checkmark & 29.52 & 0.36\\
\checkmark & \checkmark & HF (1/16) & \checkmark & \checkmark & \checkmark & \checkmark & 29.65 & 0.36\\
\checkmark & \checkmark & LF (1/16) & \checkmark & \checkmark & \checkmark & \checkmark & 29.31 & 0.36\\
\midrule
\checkmark & \checkmark & \multicolumn{5}{c|}{Layer 1 $\sim$ Layer 8} & 29.69 & 0.33\\
\checkmark & \checkmark & \multicolumn{5}{c|}{Layer 9 $\sim$ Layer 16} & 30.01 & 0.33\\
\bottomrule[0.15em]
\end{tabular}
}
\end{table*}

\begin{figure*}[t]
\begin{center}
    \includegraphics[width=1\linewidth]{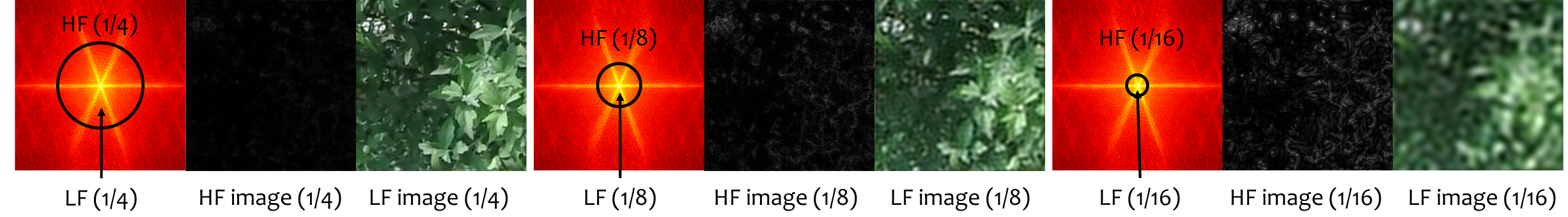}
\end{center}
\caption{Illustration on differentiating high- and low-frequencies. Although the operation of keeping high- or low-frequency is applied on features, corresponding images are shown for reference. Best viewed electronically, zoom in.} 
\label{fig:highlowfreq}
\end{figure*}

\begin{figure}[t]
\begin{center}
    \includegraphics[width=1\linewidth]{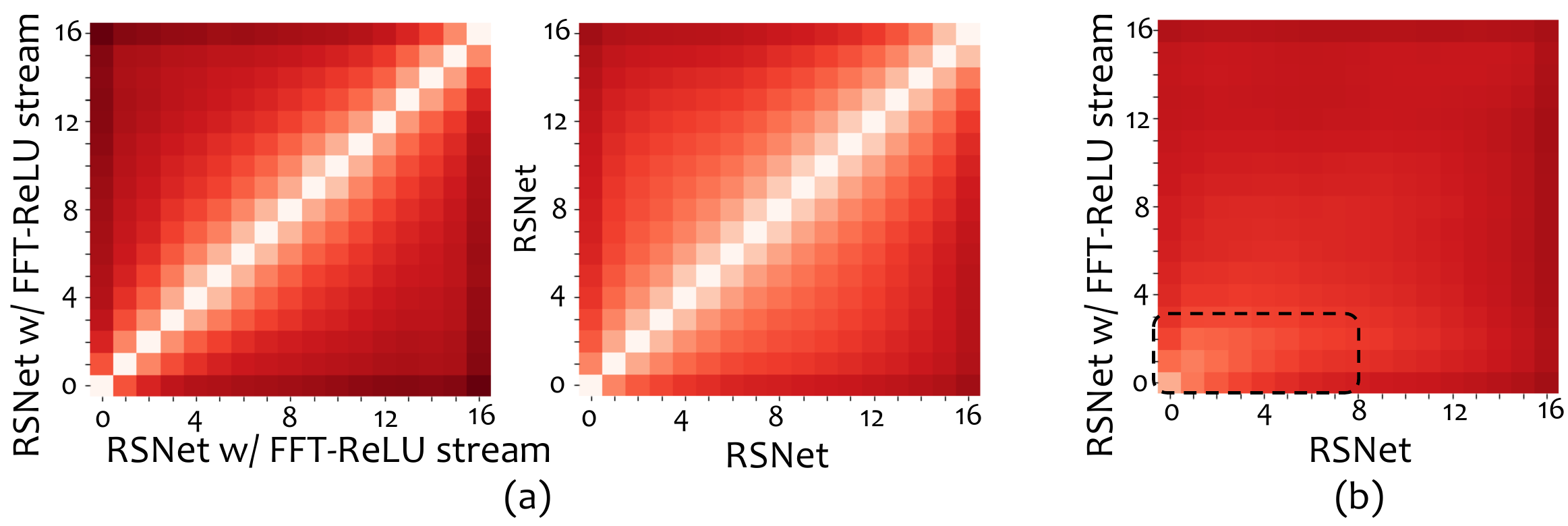}
\end{center}
\caption{CCA similarity of features between different layer pairs (a) inside RSNet w/ FFT-ReLU stream and RSNet, and (b) in RSNet w/ FFT-ReLU stream vs. RSNet. Cropped region indicates higher similarity scores.} 
\label{fig:fea_sim}
\end{figure}

\subsubsection{Low/High Frequency and Necessity of Multi-layer Res FFT-ReLU Block}
To have a better understanding on whether high- or low-frequency plays a more important role in Conv-ReLU-Conv operation in FFT-ReLU stream, we set different thresholds on differentiating high- or low- frequencies, as shown in Fig.~\ref{fig:highlowfreq}. Results are shown on the middle part in Table~\ref{tab:frequencymultilayer}. It is interesting to see that LF (1/4) works better than RSNet w/ FFT-ReLU (30.44 vs. 30.30). This is because after FFT, almost all informative frequencies lie inside LF (1/4), where HF (1/4) may contain noise frequencies, though applying ReLU on HF (1/4) can still acquire global context. For HF (1/8), LF (1/8), HF (1/16) and LF (1/16), results are slightly worse than our proposed RSNet w/ FFT-ReLU, but better than RSNet. These experiments show that all high- and low- frequencies are important in restoring sharp images.

Although our FFT-ReLU stream can learn global context, multi-layer Res FFT-ReLU Block is necessary. Multi-layer can be considered as progressively deblurring the output of the previous layers. Adding FFT-ReLU stream in the first 8 layers and the last 8 layers in RSNet obtain different deblurring results (29.69 vs. 30.01), as shown in Table~\ref{tab:frequencymultilayer}. If adding FFT-ReLU stream in the lower layers, the lack of semantic information will make the learned global information less meaningful. Besides, we show Canonical Correlation Analysis (CCA) similarities in Fig.~\ref{fig:fea_sim} (a) between layers of features inside RSNet w/ FFT-ReLU stream, and RSNet. We can see the similarity scores between low and high layers inside RSNet w/ FFT-ReLU stream are very small, which are smaller than those in RSNet. It is also interesting to observe relatively high similarity scores between lower layers of RSNet w/ FFT-ReLU stream vs. lower and middle layers of RSNet, as pointed out in Fig.~\ref{fig:fea_sim} (b).

\begin{figure}[t]
\begin{center}
    \includegraphics[width=1\linewidth]{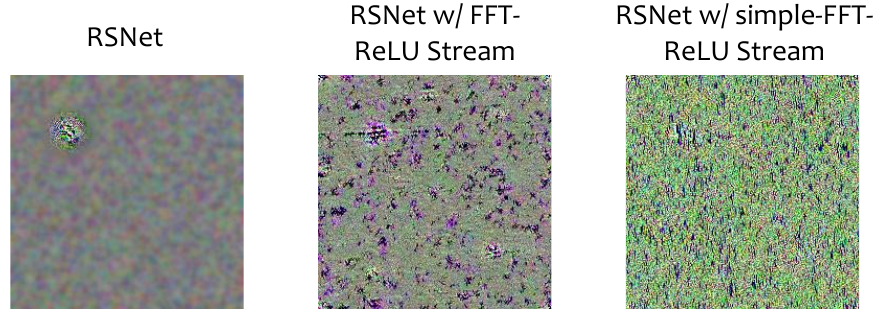}
\end{center}
\vspace{-0.8em}
\caption{Visualizations of the preferred inputs for different location units on the final output layer for RSNet, RSNet w/ FFT-ReLU Stream, and RSNet w/ simple-FFT-ReLU stream (\colorbox{gray!20}{gray} in Table 1 in the main paper).} 
\label{fig:neuron_vis_output}
\vspace{-1em}
\end{figure}

\begin{figure*}[t]
\begin{center}
    \includegraphics[width=1\linewidth]{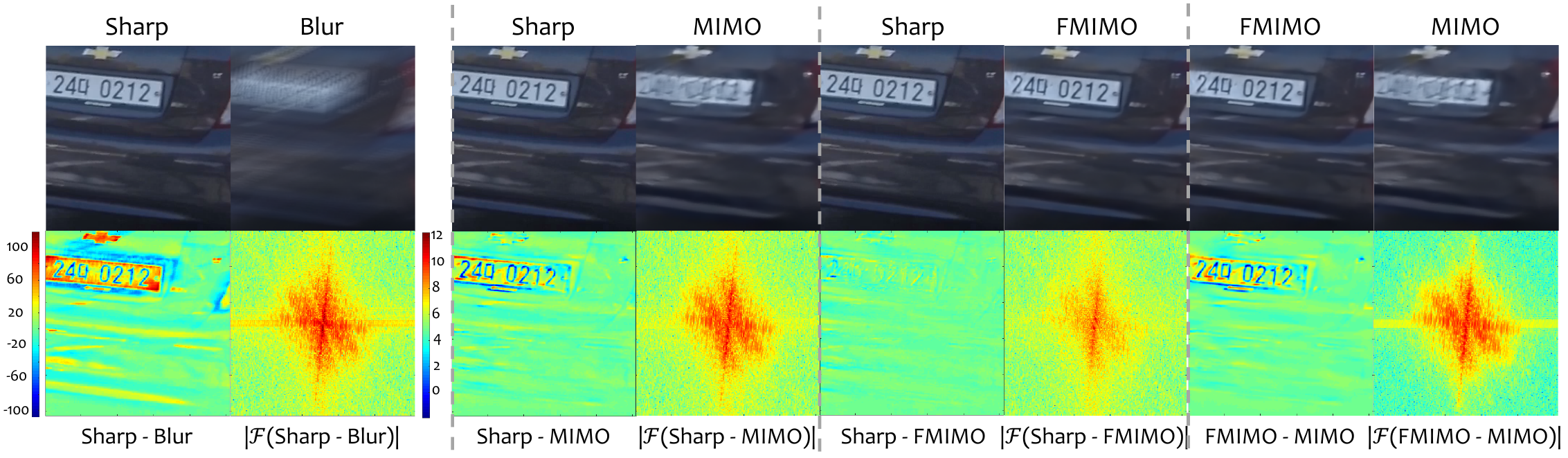}
\end{center}
\caption{Visualizations of restored image by MIMO-UNet, FMIMO-UNet and the FFT of their difference. Sharp-Blur denotes we directly subtract Blur from Sharp images, and show one channel (Green channel). The same
applies for others. $|\mathcal{F}(\cdot)|$ denotes the log of the magnitudes of 2-dimensional discrete Fourier transform, where the log operation is taking for visualization purpose.}
\label{fig:mimo-fmimo-vis}
\end{figure*}

\begin{figure}[t]
\begin{center}
    \includegraphics[width=1\linewidth]{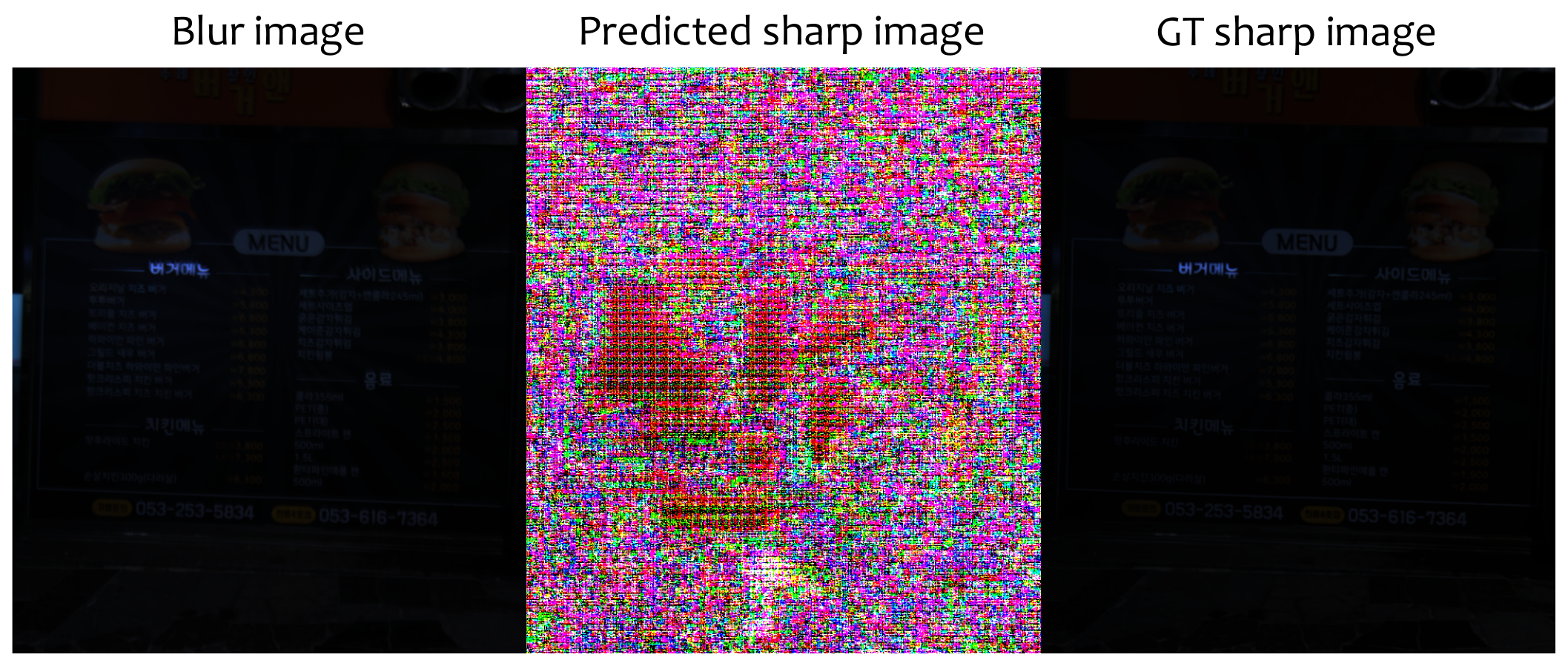}
\end{center}
\caption{Failure cases in NAFNet on RealBlur dataset.} 
\label{fig:error}
\end{figure}

\subsubsection{Other Details for Visualizing Neurons}
We provide some details for Visualizations of Neurons in Sec.~5.2 in the main paper. For training details, we follow the default setting \cite{uozbulak_pytorch_vis_2021}, which includes the maximum iterations (150), optimizer (SGD), initial learning rate (6), weight decay (0.0001), and the Gaussian filter which is applied every 4 iterations. We also visualize the neuron in the output layer of different networks, including RSNet w/ simple-FFT-ReLU. For the output layer, we adopt a slightly different strategy. \textit{I.e.}, instead of maximizing a selected neuron, we first apply Sobel filter on the output image to obtain the edge map of the image, then we learn the best input image by maximizing the neuron on location $[\frac{1}{4}H, \frac{1}{4}W]$ on the obtained edge map. We propose to visualize the output neuron after applying a Sobel filter, because the ability of maximizing an edge point is related to the ability of image deblurring. Visualizations of the preferred inputs for different location units on the final output layer are shown in Fig.~\ref{fig:neuron_vis_output}. It is easy to see that RSNet can only learn local context, while the other two networks have the ability to capture global context.

\subsubsection{Visualizations of MIMO-UNet w/ and w/o Res FFT-ReLU Block}
We denote the MIMO-UNet with our proposed Res FFT-ReLU Block as FMIMO-UNet. Since end-to-end image deblurring architecture learns the discrepancy between blur and sharp image pairs, given a deblurring architecture, it is intuitive to visualize the difference between the restored sharp image with the ground-truth sharp image. This difference indicates how far to go for the model to restore an ideal sharp image. We also take the FFT of the difference, as shown in Fig.~\ref{fig:mimo-fmimo-vis}. Compared with MIMO-UNet, by using our Res FFT-ReLU Block, FMIMO-UNet can compensate more frequency information (see $|\mathcal{F}(\text{FMIMO}-\text{MIMO})|$, and comparisons between $|\mathcal{F}(\text{Sharp}-\text{FMIMO})|$ and $|\mathcal{F}(\text{Sharp}-\text{MIMO})|$).

\begin{table}[t]
\renewcommand\arraystretch{1}
\footnotesize
\centering
\caption{Parameters, FLOPs and runtime
(average testing time per image in GoPro dataset on a NVIDIA 3090 GPU) comparison. {\color{red}${\ddagger}$} means testing an image for 4 times by test time augmentation.}
\label{tab:flops}
\resizebox{1\linewidth}{!}{
\begin{tabular}{lc|ccc}
\toprule[0.15em]
Model& PSNR & Params.& FLOPs & Runtime\\
& & (M) & (G) & (s)\\
\midrule[0.09em]
DMPHN \cite{Zhang2019deep} & 31.20 & 21.7 & - & 0.307\\
DBGAN \cite{Zhang2020deblurring} & 31.10 & 11.6 & 759.85 & 1.298\\
MPRNet \cite{Zamir2021multi} & 32.66 & 20.1 & 777.01 & 1.002\\
MIMO-UNet \cite{Cho2021rethinking} & 31.37 & 6.8 & 67.17 & 0.153\\
{MIMO-UNet+} & 32.45 & 16.1 & 154.41 & 0.309\\
{MIMO-UNet++} & 32.68 & 16.1{\color{red}${\ddagger}$} & 617.63 & 1.185\\
SDWNet \cite{Zou2021sdwnet} & 31.36 &7.2 & 189.68 & 0.533\\
UFormer \cite{Wang2022uformer} & 33.06 & 50.9 & 89.46 & 0.442\\
Restormer \cite{Zamir2021restormer} & 32.92 & 26.1 & 141.00 & 1.14\\
NAFNet32 \cite{Chen2022simple} & 32.85 & 17.1 & 16.00 & 0.132\\
NAFNet64 \cite{Chen2022simple} & 33.69 & 65.0 & 63.50 & 0.329\\
\midrule
FMIMO-UNet-small & 32.52 & 4.4 & 44.60 & 0.244\\
FMIMO-UNet & {33.08} & 8.2 & 80.21 & 0.339\\
FMIMO-UNet+ & {33.52} & 19.5 & 187.04 & 0.948\\
FNAFNet32 & 33.12 & 17.8 & 23.10 & 0.185\\
FNAFNet64 & \textbf{33.85} & 68.0 & 72.40 & 0.366\\
\bottomrule[0.15em]
\end{tabular}
}
\end{table}

\subsubsection{Evaluation of FMIMO-UNet and FNAFNet}
\paragraph{Detailed Analysis for Evaluation of FMIMO-UNet and FNAFNet}
The results of FMIMO-UNet-small, FMIMO-UNet, and FMIMO-UNet+ are reported after cropping the input image to the training size of 256 with an overlap size of 32, \emph{i.e.}, sliding window strategy with window size 256 and stride 224. This further improves the performance for FMIMO-UNet from 32.71 dB to 33.08 dB. As shown in Table 5 in the main paper, our FMIMO-UNet-small outperforms MIMO-UNet by 0.79 dB PSNR with 2.4M less parameters (see Table~\ref{tab:flops}). As mentioned above, because of our limited computation resource, for FNAFNet32, we add Res FFT-ReLU block to the first three encoder blocks and their skip connected decoder blocks. For FNAFNet64, we only add Res FFT-ReLU block to the first encoder block and its skip connected decoder block. We set the batch size to 32 for both FNAFNet32 and
FNAFNet64. But for NAFNet64, the batch size was set as 64 in the paper \cite{Chen2022simple}. FNAFNet32 outperforms NAFNet32 by 0.27 dB and FNAFNet64 outperforms NAFNet64 by 0.16 dB. We do not report the results on NAFNet64/FNAFNet64 for RealBlur dataset, because some predicted sharp images look weird for NAFNet64, as shown in Fig.~\ref{fig:error}. This may due to the reason that the model cannot fit the distribution of the sharp image.

\paragraph{Parameters, FLOPs and Runtime Comparison}
We compare parameters, FLOPs and runtime of our FMIMO-UNet and FNAFNet with the state-of-the-arts (see Table~\ref{tab:flops}). FLOPs number is calculated using ptflops \footnote{https://github.com/sovrasov/flops-counter.pytorch} with the input size of 256$\times$256 \cite{Zou2021sdwnet}. Runtime is measured by using the released test code of each method to run the entire GoPro \cite{Nah2017deep} testing dataset, and obtaining the average runtime (second) per image on our environment with one NVidia 3090 GPU. FMIMO-UNet+ takes much less FLOPs than MIMO-UNet++ and MPRNet.

\subsection{Qualitative Comparisons of FMIMO-UNet with Other Competitors}

\begin{figure*}[h]
\begin{center}
    \includegraphics[width=0.9\linewidth]{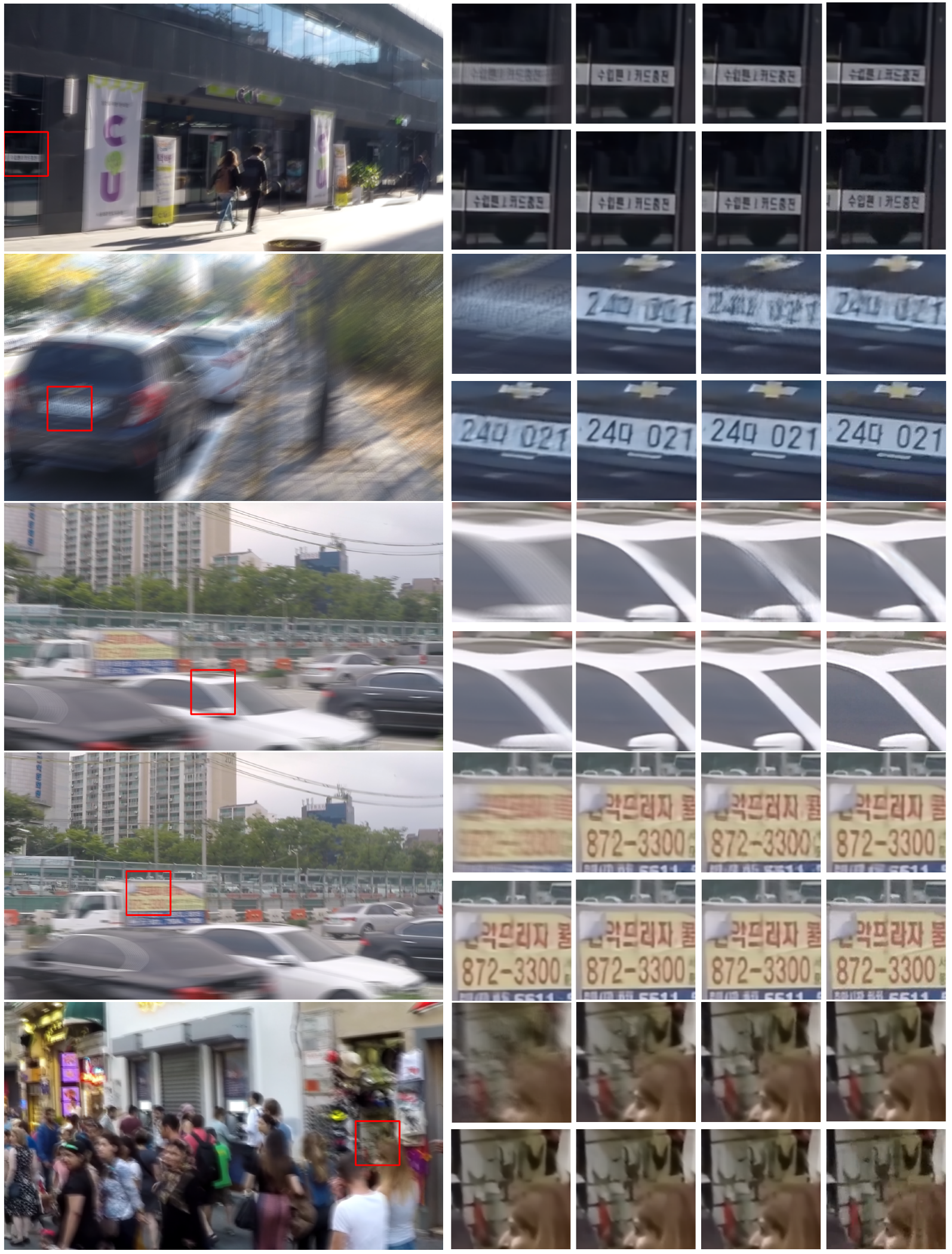}
\end{center}
\caption{Examples on the GoPro test dataset. The original blur image and the zoom-in patches are shown. From left-top to right-bottom are blurry image, results obtained through, MIMO-UNet+ \cite{Cho2021rethinking}, HINet \cite{Chen2021hinet}, MPRNet \cite{Zamir2021multi}, Restormer \cite{Zamir2021restormer}, FMIMO-UNet, FMIMO-UNet+, and the groundtruth sharp image.} 
\label{fig:gopro}
\vspace{-0.8em}
\end{figure*}

\begin{figure*}[h]
\begin{center}
    \includegraphics[width=0.75\linewidth]{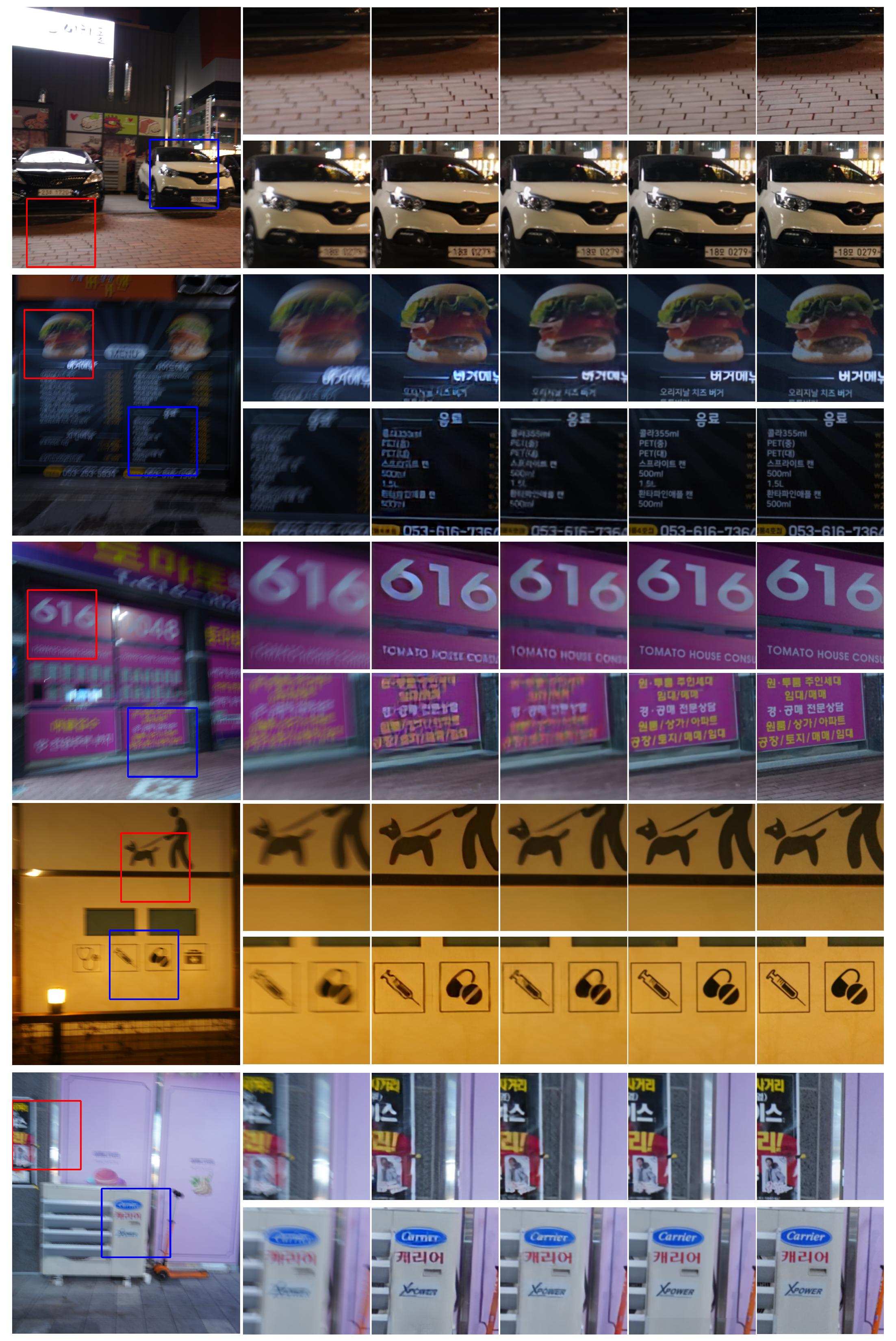}
\end{center}
\caption{Examples on RealBlur datasets. The original blur image and the zoom-in patches are shown. From left to right are blurry image, results obtained through DeblurGAN-v2 \cite{Kupyn2019deblurgan}, SRN \cite{Tao2018scale}, FMIMO-UNet+ and the groundtruth sharp image.} 
\label{fig:realblur}
\end{figure*}

\begin{figure*}[h]
\begin{center}
    \includegraphics[width=1\linewidth]{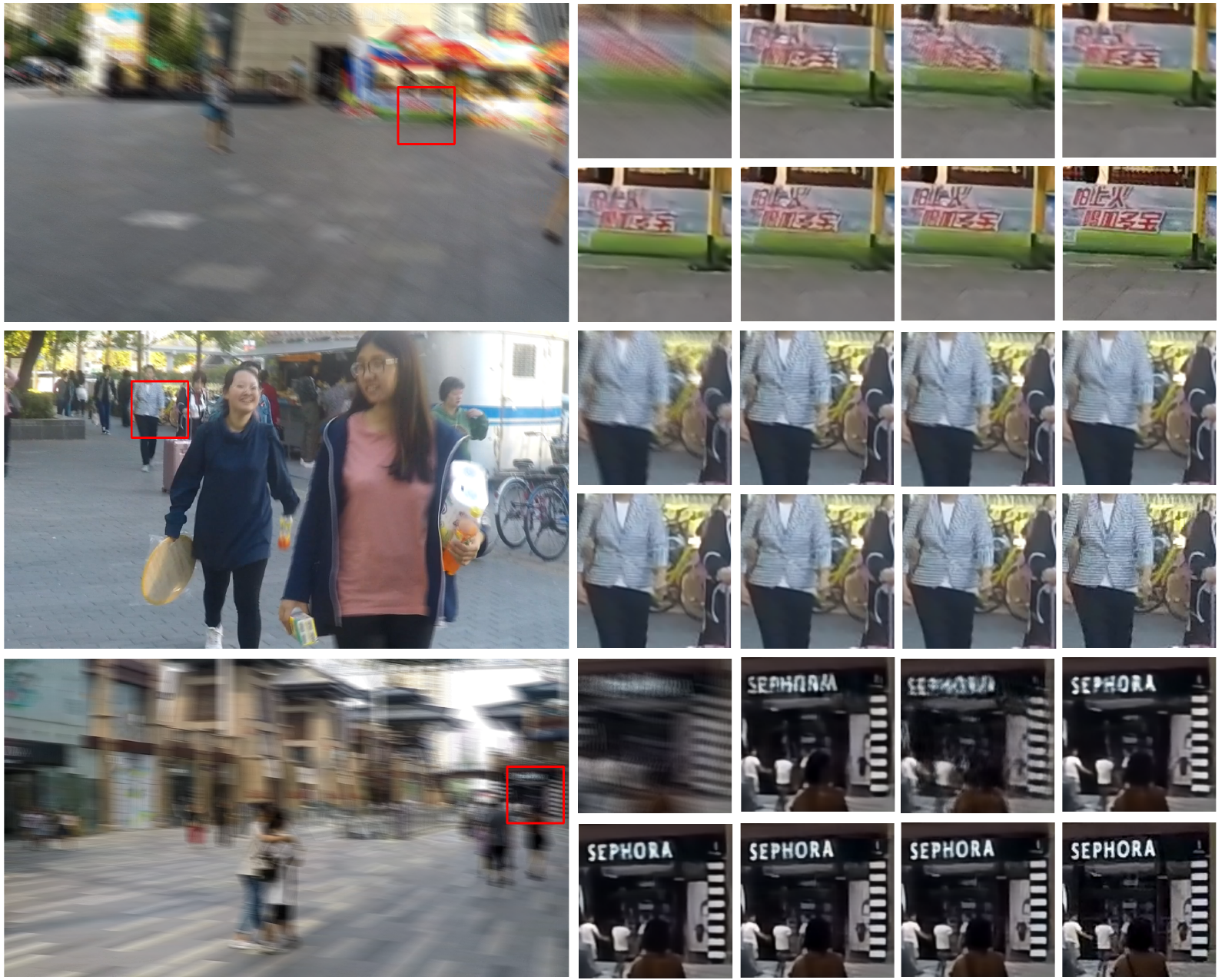}
\end{center}
\caption{Examples on HIDE datasets. The original blur image and the zoom-in patches are shown. From left-top to right-bottom are blurry image, results obtained through MIMO-UNet+ \cite{Cho2021rethinking}, HINet \cite{Chen2021hinet}, MPRNet \cite{Zamir2021multi}, Restormer \cite{Zamir2021restormer}, FMIMO-UNet, FMIMO-UNet+ and the groundtruth sharp image.} 
\label{fig:hide}
\end{figure*}

Some of the predicted sharp images from the GoPro, HIDE, and RealBlur datasets are shown in Fig.~\ref{fig:gopro}, Fig.~\ref{fig:hide} Fig.~\ref{fig:realblur}, respectively. Results from competitors are tested by using their released models. FMIMO-UNet is more successful in deblurring local details and structures compared with others.

\end{document}